\documentclass[twocolumn,preprint]{elsarticle}

\usepackage{lineno}
\usepackage{url}
\usepackage{hyperref}
\usepackage{times}
\usepackage{epsfig}
\usepackage{graphicx}
\usepackage{amsmath}
\usepackage{multirow}
\usepackage{amssymb}
\usepackage{subfig}
\usepackage{graphicx}
\usepackage{float}
\usepackage{booktabs}
\usepackage{textcomp}
\journal{Knowledge-based Systems}

\begin{document}

\twocolumn[{
\begin{frontmatter}

\title{SAIA: Split Artificial Intelligence Architecture for Mobile Healthcare Systems}
% \tnotetext[mytitlenote]{Fully documented templates are available in the elsarticle package on \href{http://www.ctan.org/tex-archive/macros/latex/contrib/elsarticle}{CTAN}.}

\author{Di Zhuang, Nam Nguyen, Keyu Chen, and J. Morris Chang\\
{\tt\small \{dizhuang, namnguyen2, keyu, chang5\}@usf.edu}\\}
\address{Department of Electrical Engineering, University of South Florida, Tampa, FL 33620}
% \fntext[myfootnote]{Since 1880.}

% \author[mymainaddress,mysecondaryaddress]{Elsevier Inc}

% \author[mysecondaryaddress]{Global Customer Service\corref{mycorrespondingauthor}}
% \cortext[mycorrespondingauthor]{Corresponding author}

% \address[mymainaddress]{1600 John F Kennedy Boulevard, Philadelphia}
% \address[mysecondaryaddress]{360 Park Avenue South, New York}

\begin{abstract}

As the advancement of deep learning (DL), the Internet of Things and cloud computing techniques for biomedical and healthcare problems, mobile healthcare systems have received unprecedented attention. Since DL techniques usually require enormous amount of computation, most of them cannot be directly deployed on the resource-constrained mobile and IoT devices. Hence, most of the mobile healthcare systems leverage the cloud computing infrastructure, where the data collected by the mobile and IoT devices would be transmitted to the cloud computing platforms for analysis. However, in the contested environments, relying on the cloud might not be practical at all times. For instance, the satellite communication might be denied or disrupted. We propose SAIA, a Split Artificial Intelligence Architecture for mobile healthcare systems. Unlike traditional approaches for artificial intelligence (AI) which solely exploits the computational power of the cloud server, SAIA could not only relies on the cloud computing infrastructure while the wireless communication is available, but also utilizes the lightweight AI solutions that work locally on the client side, hence, it can work even when the communication is impeded. In SAIA, we propose a meta-information based decision unit, that could tune whether a sample captured by the client should be operated by the embedded AI (i.e., keeping on the client) or the networked AI (i.e., sending to the server), under different conditions. In our experimental evaluation, extensive experiments have been conducted on two popular healthcare datasets. Our results show that SAIA consistently outperforms its baselines in terms of both effectiveness and efficiency.

\end{abstract}

\begin{keyword}
\texttt{Split Artificial Intelligence; Mobile Healthcare System; Internet of Things; Algorithm Selection; Deep Learning; Machine Learning; Fusion; Skin Lesion; Nail Fungus; Onychomycosis; Embedded AI; Networked AI; Decision Unit; Data Pre-processing; Resource-constrained.}
\end{keyword}
\end{frontmatter}
}]

% \linenumbers

\section{Introduction}\label{sec:introduction}
As the advancement of modern technologies, such as wireless communication, data mining, machine learning, the Internet of Things (IoT), cloud computing and edge computing, the mobile healthcare systems become more and more feasible and popular. Numerous intelligent mobile healthcare systems are developed on various mobile and IoT devices \cite{farahani2020healthcare}. The emergence and breakthrough of deep learning, that has been shown to achieve extraordinary results in a variety of real-world applications, such as skin lesion analysis \cite{perez2019solo}, active authentication \cite{wu2016cost}, facial recognition \cite{nguyen2019autogan, zhuang2017fripal}, botnet detection \cite{zhuang2017peerhunter, zhuang2018enhanced} and community detection \cite{zhuang2019dynamo}, is one of the primary driver for such mobile healthcare systems. However, since the deep learning techniques require enormous amount of computation resources, most of them cannot be directly deployed on the %computation-constrained and energy-limited
resource-constrained mobile and IoT devices.

One common solution to tackle such problem is cloud computing, where the data could be transmitted to the cloud computing platforms for operations.
For instance, several Machine Learning as a Service (MLaS) systems were introduced in the recent years (e.g., Google Cloud AutoML \cite{AutoML} and Amazon SageMaker \cite{SageMaker}). These systems were mostly intended to utilize the high computational power of cloud servers, for ML applications, in addition to enable scalability in the cloud (horizontal scaling). However, in contested environments, relying on the server to generate actionable intelligence might not be practical at all times. For instance, the satellite communication might be denied or disrupted. For such situations, the mobile and IoT devices have to be enabled to generate actionable intelligence that might be required for the success of certain operations (i.e., providing healthcare services). Hence, it is imperative to design a Split Artificial Intelligence Architecture (SAIA), unlike the traditional AI architecture, that can not only exploit the computational power of the server, but also utilize lightweight AI solutions that work locally on the mobile or IoT devices.

Designing an effective and efficient SAIA system has to meet several challenging requirements.
First, the client side (i.e., mobile or IoT devices) should have lightweight (in terms of storage size, power consumption and inference time) AI solutions, that could provide fundamental services (i.e., acceptable classification precision for certain classes or subsets of data) even when the satellite communication is denied or disrupted.
Second, the server side (i.e., cloud server) should have complex ``full-sized'' AI solutions, that could provide the state-of-the-art performance on the selected applications.
Third, the usage of AI solutions (on the whole or subset of the data) shifting between the client side and the server side should depend on the application precision requirement, the resource availability and the data characteristics, and such trade-off should be able to be optimized.
Last but not least, the adjustment of AI usage between the client and the server should be efficient and intelligent. For instance, if the lightweight AI is able to recognize the class of given data (w.h.p.), the data should not be sent to the server, even when the communication is unimpeded.

To date, a few approaches have been proposed to tackle the problem of running deep learning techniques on the mobile and IoT devices. For instance, Knowledge  Distillation (KD) \cite{hinton2015distilling, ba2014deep, polino2018model} has been proposed to compress a model by teaching a simplified student DNN model, step by step, exactly what to do using a complex pre-trained teacher DNN model, and then deploy the student DNN model on the mobile devices \cite{wang2019private}. Although KD could dramatically reduce the complexity of the student model, the overall performance of a student model still would be as good as its teacher model. Moreover, solely deploying a lightweight model on the client side loses the chance and advantage of using a more advanced model on the server side, that could be the ensemble/fusion of several well-trained DNN models.
Split-DNN architectures \cite{jeong2018computation, kang2017neurosurgeon, lane2016deepx} have also been proposed to offload the execution of complex DNN models to compute-capable servers from the mobile and IoT devices, where a DNN is split into head and tail sections, deployed at the client side and the server side, respectively. Matsubara et al. \cite{matsubara2019distilled} proposes a KD-based Split-DNN framework to reduce the communication cost between the client and the server. However, such approaches usually cannot fully rely on the client-side model, thus unable to work if the communication is impeded. To summarize, KD and split-DNN focus on either deploying lightweight models on the client side or pushing the most of the DNN computation to the server side in an efficient fashion. However, none of the existing approaches could adjust the AI usage on between the client and the server depending on the device's condition (e.g., storage  size,  power  consumption and  communication bandwidth).

In this paper, we propose SAIA, a Split Artificial Intelligence Architecture for mobile healthcare systems.
SAIA enables the client to produce actionable intelligence locally using its embedded AI unit (e.g., conventional ML classifiers). When the satellite communication is available, the reduced feature data (or compressed raw data) could be uploaded to the server, and be processed by the networked AI units that utilize more powerful AI algorithms (e.g., ensemble of multiple advanced DNN classifiers); thus generating more confident and detailed AI results.
The embedded AI client might need communication with the server if the confidence score of the decision is below a certain threshold, or periodically, when the satellite communication is available, to generate more confident and detailed AI results using the more powerful networked AI on the server side.
In SAIA, we also propose a decision unit that trains on the meta-information (e.g., soft labels) outputted by the embedded AI and is deployed on the client side to decide and control whether a sample captured by the client should be operated on the client side or sent to the server side. We also enable the decision unit to utilize a parameter, namely $\epsilon$, to tune the criteria of how much data could be sent to the networked AI. As such, our SAIA framework could work under different conditions (e.g., unimpeded communication bandwidth or satellite communication is denied or disrupted).

In the experimental evaluation, we trained three conventional machine learning models (i.e., SVM, RF and DART) for the embedded AI and an ensemble of twelve advanced DNN models for the networked AI, on two popular healthcare benchmark datasets: the ISIC research dataset for skin image analysis \cite{gutman2016skin, codella2018skin, codella2019skin} and the onychomycosis (a.k.a. Nail fungus) dataset \cite{han2018deep}.
Our experimental results show that our SAIA framework is effective and efficient while switching the computation between the embedded AI and the networked AI. Also, our design of SAIA's decision unit consistently outperforms its baseline (i.e., randomly selected sending) in terms of both effectiveness and efficiency.

To summarize, our work has the following contributions:

$\bullet$ We present SAIA, a novel, effective and efficient split artificial intelligence architecture. To the best of our knowledge, this is the first work to apply split artificial intelligence architecture in mobile healthcare systems.

$\bullet$ In SAIA, we propose a meta-information based decision unit, that could tune whether a sample captured by the client should be operated by the embedded AI or the networked AI, under different conditions.

$\bullet$ A comprehensive experimental evaluation on two large scale healthcare datasets has been conducted. We have implemented three popular conventional MLs as the embedded AI, and utilized an ensemble of twelve advanced DNN classifiers as the networked AI. For the sake of reproducibility and convenience of future studies about split artificial intelligence architecture, we have released our prototype implementation of SAIA, information regarding the experiment datasets and the code of our evaluation experiments.\footnote[1]{\url{https://tinyurl.com/y92epzfd}}

The rest of this paper is organized as follows:
Section~\ref{sec:methodology} presents SAIA, including the design of the embedded AI, the networked AI and the decision unit.
Section~\ref{sec:experimentalEvaluation} presents the experimental evaluation.
Section~\ref{sec:relatedWork} presents the related literature review.
Section~\ref{sec:conclusion} concludes.

\begin{figure*}[!h]
  \centering
  \includegraphics[width=1.0\linewidth]{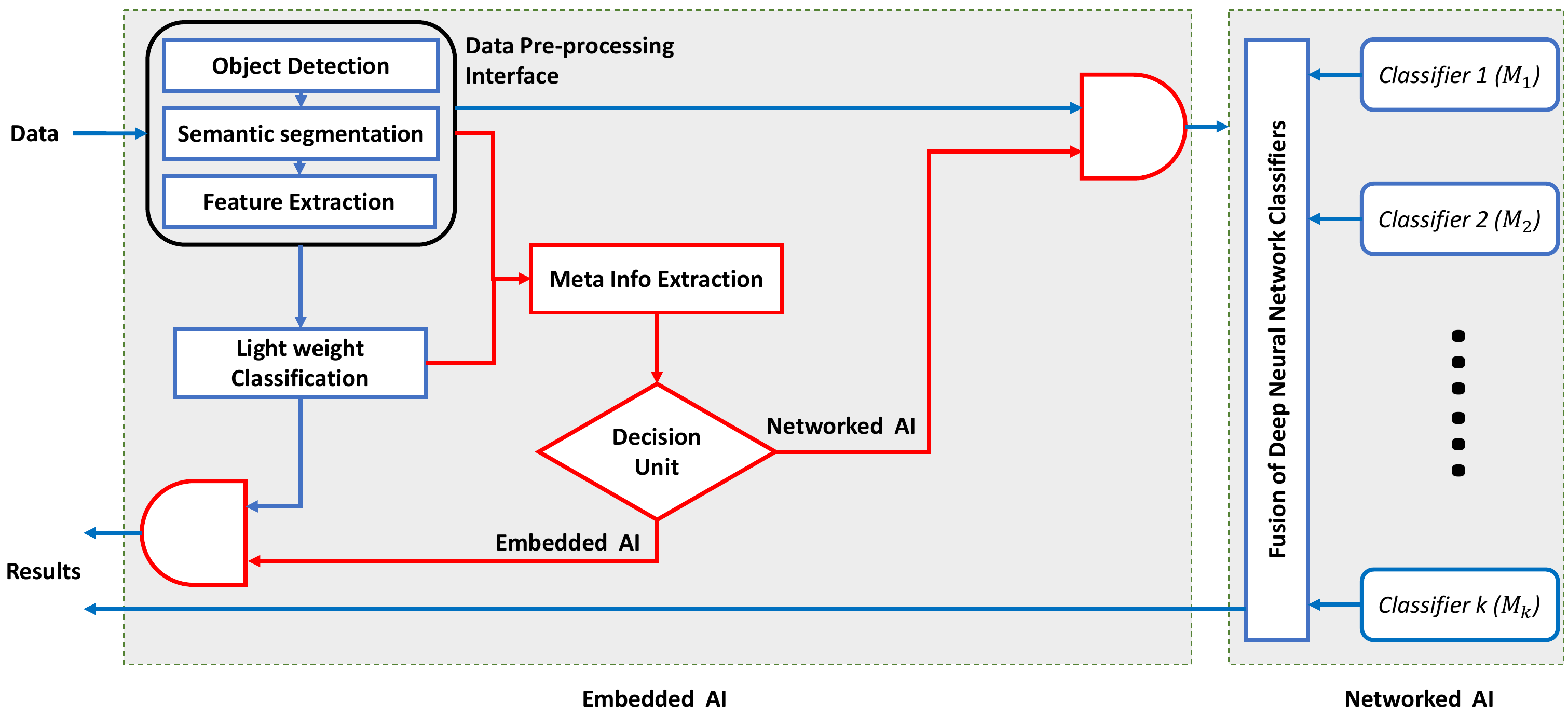}
  \caption{The Overview of SAIA Framework.}
  \label{Overview}
\end{figure*}

\section{Methodology} \label{sec:methodology}

\subsection{SAIA Framework Overview} \label{sec:methodology_overview}
Our proposed Split Artificial Intelligence Architecture (SAIA), as shown in Fig.~\ref{Overview}, consists of four components (i.e., the data pre-processing interface, the embedded AI, the networked AI and the decision unit) that work synergistically between the client side and the server side.
For each use case, SAIA has two phases: preparation and operation.
In the preparation phase, four components would be prepared and trained accordingly: (i) the data pre-processing interface (including objection detection, semantic segmentation and feature extraction) (Section~\ref{sec:methodology_DPI}); (ii) the embedded AI contains certain lightweight classification classifier(s) (Section~\ref{sec:methodology_embedded}); (iii) the networked AI trains a multi-classifier fusion of several advanced DNN classifiers (Section~\ref{sec:methodology_networked}); and (iv) the decision unit is a lightweight ML classifier that trains on a set of labeled meta data (Section~\ref{sec:methodology_du}).

In the operation phase, (i) the client (i.e., mobile or IoT devices) receives the data, passes the data through the data pre-processing interface (including objection detection, semantic segmentation and feature extraction); (ii) the client evaluates the data on the embedded AI (i.e., the lightweight classifier(s)), and produces the unlabeled meta data accordingly; (iii) the decision unit (DU) evaluates the meta data, if DU decides to keep the data on the client side, the testing result of the embedded AI is returned, otherwise, the data would be sent to the networked AI for further evaluation. Below presents the details about the design of each components.

\subsection{Data Pre-processing Interface} \label{sec:methodology_DPI}
In this component, we design and implement a set of objection detection, semantic segmentation and feature extraction algorithms that could fit on various image-based healthcare applications.

{\bf Objection detection:} Since the medical images captured by the mobile and IoT devices usually contain complex background, it is of vital importance to separate the region-of-interest (ROI) from the background. We investigate two fast object detection approaches: (i) Faster R-CNN \cite{ren2015faster} and (ii) Single Shot Detector (SSD) \cite{liu2016ssd}. We also conducted a preliminary experiment using both object detection approaches on the onychomycosis dataset \cite{han2018deep} (more details about the dataset are described in Section~\ref{sec:experimentalEvaluation_data}), where we annotated 2,000 images that each contains a full-hand, and then trained and applied Faster R-CNN %\cite{ren2015faster}
and SSD %\cite{liu2016ssd}
on the annotated images. In the result, Faster R-CNN %\cite{ren2015faster}
obtains in a Jaccard Index %\cite{real1996probabilistic}
of 99.6 in comparison with 98.2 obtained from SSD. %\cite{liu2016ssd}.
Therefore, we decided to apply Faster R-CNN on the onychomycosis dataset (Section~\ref{sec:experimentalEvaluation_data}).

{\bf Semantic segmentation:} % After applying conventional data augmentation, which are rotating, flipping and all combination of rotating and flipping,
We utilize Otsu's thresholding segmentation algorithm \cite{zhang2008image} for the image semantic segmentation. By finding the optimal threshold from the histogram of pixel counts, the algorithm isolates the region-of-interest (ROI) (e.g., objects) from the complex backgrounds. For instance, it could separate the skin lesion from the normal skin and artifacts (hairs, badges and black borders). Compared with the other semantic segmentation approaches, such as U-Net CNN \cite{ronneberger2015u}, Otsu's thresholding segmentation algorithm not only has been shown to be effective in many medical image segmentation tasks \cite{xiao2018weighted, zhao2018lung, li2018h}, but also is more efficient in terms of storage usage, energy consumption and inference time while deploying on mobile and IoT devices.

{\bf Feature extraction:} We design and apply different sets of feature extraction techniques to different healthcare applications. For instance, in skin lesion detection, melanoma (i.e., cancerous skin lesion) usually proliferates asymmetrically then appears as irregular shapes. Hence, the derived segmentation maps and corresponding gray-scale images are used to extract 9 structural features, including: asymmetric index \cite{abuzaghleh2014automated}, eccentricity, perimeter, max/min/mean intensity, solidity, compactness and circularity \cite{sancen2018quantitative} . Furthermore, color variations are also effective characteristics to distinguish different types of skin lesions. Instead of using convention RGB color space, we employ CIELUV color space that enables us to well perceive the differences in colors. Besides, LUV color space also decouples the chromaticity (UV) and luminance (L), which yields invariant features in respect to the light condition. Instead of taking only statistics (mean, standard deviation, skewness and kurtosis) from LUV histogram \cite{seeja2019deep}, we utilize the whole distribution of colors , which are separated into 3 channels (i.e., L, U and V). As a result, $3 \times 255$ color features would be generated from three normalized histograms. Moreover, we observe that the texture of lesions can also distinguish skin lesion types. Hence, local binary patterns (LBP) \cite{ahonen2006face} analysis is applied to capture the textured information. We investigated several sets of radius and number of surrounding points then observed that using radius of 3 and 8 neighboring points yields the best performance of Embedded-AI with 26 textured features from each normalized LBP histogram. On the other hand, since the onychomycosis detection does not appear to be ``shape-sensitive'', we only adopt the color-based features (i.e., LUV) and texture features (i.e., LBP) to it.

\subsection{Client-side Embedded AI} \label{sec:methodology_embedded}
Since the client device has limited computational power, and limited battery life, we intend to utilize lightweight ML algorithms. The embedded AI solutions will be used to generate initial classification results; thus enabling embedded artificial intelligence. The use of lightweight algorithms would decrease the burden on battery life; which enables the operator's equipment to last longer in contested environments. These algorithms also require less computational power, which results in producing intelligence in a more timely manner (than more complex algorithms).

These lightweight classification algorithms could be distance-based algorithms that are based on Euclidean or Manhattan distances, or it could be logistic regression. In both cases, the computation would be linear in the number of (reduced) features; and do not involve multiple layers of computations; thus, providing more timely intelligent results and consuming less power. Examples of such algorithms include:

{\bf Decision Tree (DT):} DT \cite{quinlan1986induction} is a non-linear classifier. It is rule-based learning method that would construct a tree where the leaves represent class labels, and the branches represent conjunctions of features that lead to those
class labels. The tree structure would depend on the algorithm and data used to generate it, but in certain situations, it might be lightweight and suitable for our embedded AI (e.g. if it was of linear complexity). The advantage is that it can handle non-linearly separable data (better than Logistic Regression).

{\bf Random Forests (RF):} Decision tree classifier \cite{quinlan1986induction} usually yields high-variance and low-bias results, thus bagging (or bootstrap aggregation) is a remedy for such issues. RF \cite{liaw2002classification} is a large collection of de-correlated trees, which are aggregated by taking their averages. Besides, trees generated in bagging is identically distributed, the expected value from this bagging set of trees is the same as the expectation of any tree in this set. Thus, variance reduction is only remedy of improvement.

{\bf Support Vector Machine (SVM):} SVM \cite{suykens1999least} is a very popular non-linear classifier. It is a maximal margin classifier, meaning it tries to find a separating hyperplane that maximizes the margin between the different classes (while logistic regression, for example, tries to find any separating boundary). Using the Kernel trick, Kernel SVM can effectively discriminate between non-linearly separable classes, without incurring the cost of explicitly transforming the data to higher dimensions. A sample Kernel function is the RBF Kernel:
	\begin{equation}
	K(x, y) = exp\left(-\frac{\|x-y\|^2}{2\sigma^2}\right)
	\end{equation}

Kernel SVM requires storing a number of the training samples, called support vectors. Let this number be $n << N$, and it is much smaller than the total number of training samples $N$. This means that Kernel SVM would have space and time complexity of $O(n \cdot M)$.
While Kernel SVM is very useful in many applications, its training might not scale well with datasets that have large number of training samples (beyond tens of thousands).

{\bf Dropouts meet Multiple Additive Regression Trees (DART):} DART \cite{rashmi2015dart} is an evolution of gradient boosting machine, that adopts the dropouts for regularization (preventing over-fitting) from deep neural networks. Boosted trees with XGboost \cite{chen2016xgboost} is one of the most well-performed learning structure, which results in a great number of winning solutions of data science competitions \cite{nielsen2016tree}. Apart from classification, boosting tree can be used in a wide range of problem such as regularized regression (Ridge and Lasso) \cite{tutz2007boosting}, quantile regression \cite{fenske2011identifying} or survival analysis \cite{buhlmann2007boosting, nguyen2020gradient}. Motivated by systems optimization and fundamental principle of machine learning, XGboost \cite{chen2016xgboost} is an efficient and flexible library with implementation of parallel tree boosting which enables fast and accurate results.

In our preliminary experiments, DART \cite{rashmi2015dart} outperforms the conventional multiple additive regression tree (MART) \cite{friedman2002stochastic} and AdaBoost \cite{hastie2009multi} in both datasets (i.e., skin lesion \cite{tschandl2018ham10000, codella2018skin, combalia2019bcn20000} and onychomycosis dataset \cite{han2018deep}) in terms of both training time and accuracy. All the implementations regarding DART \cite{rashmi2015dart} in this work utilize both XGboost library for Python 3 \cite{xgboost} and Microsoft's LightGBM framework \cite{ke2017lightgbm} for gradient boosting machine.

\subsection{Server-side Networked AI} \label{sec:methodology_networked}
Since the embedded AI algorithms might not handle non-linearly separable classification problems well, we aim to use more powerful algorithms in the networked AI. Such algorithms would include various advanced DNNs, and we design it as a multi-classifier fusion of those classifiers.
These algorithms are more computationally intensive; but they can produce more accurate results; thus, providing more confident intelligence. This type of computation can be facilitated on the server side by using state-of-the-art big data technologies. Below presents the details of our multi-classifier fusion approach.

\begin{figure*}[!h]
  \centering
  \includegraphics[width=1.0\linewidth]{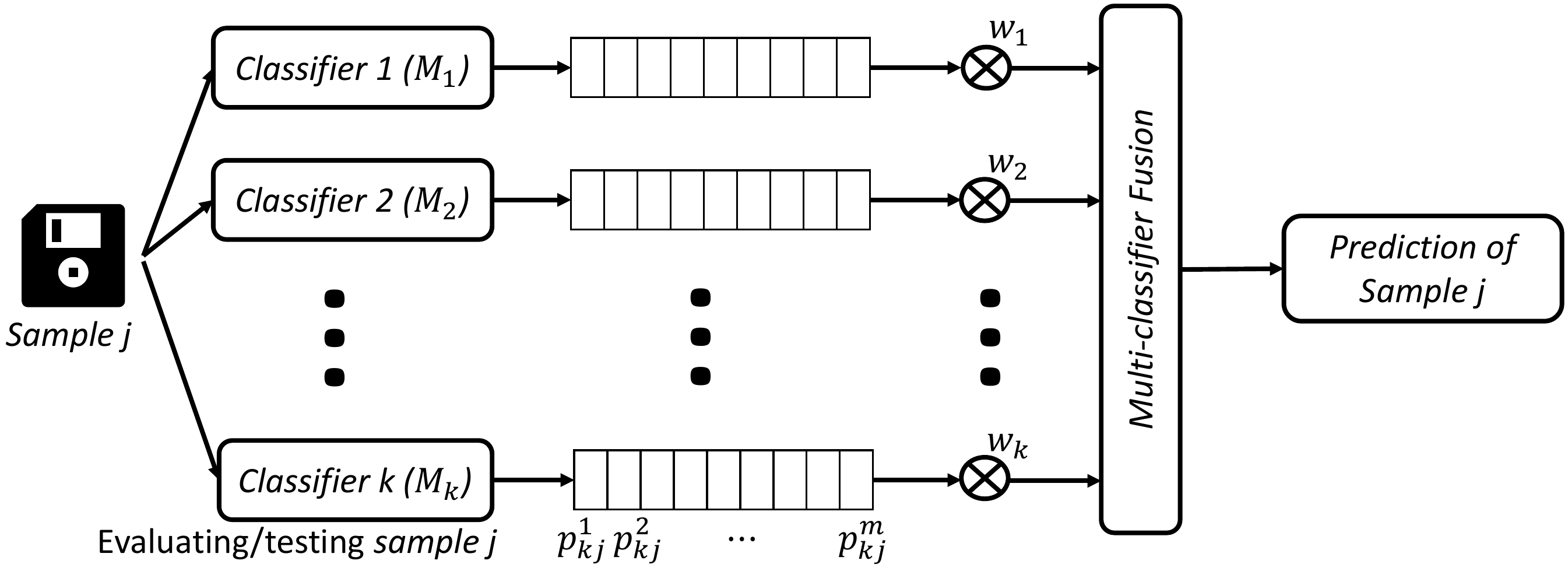}
  \caption{Networked AI: Multi-classifier Fusion.}
  \label{Networked_AI}
\end{figure*}

\subsubsection{Multi-classifier Fusion} \label{ES_Problem}
In multi-classifier fusion, we define a classification space, as shown in Figure~\ref{Networked_AI}, where there are $m$ classes and $k$ classifiers. Let $\mathcal{M}=\{M_{1}, M_{2}, \dots, M_{k}\}$ denote the set of base classifiers and $\mathcal{C}=\{C_{1}, C_{2}, \dots, C_{m}\}$ denote the set of classes. Let $p^{m}_{kj}$ denote the posterior probability of given sample $j$ identified by classifier $M_{k}$ as belonging to class $C_{m}$, where $P_{kj}=\{p^{1}_{kj}, p^{2}_{kj}, \dots, p^{m}_{kj}\}$ and $\sum_{l=1}^{m} p^{l}_{kj} = 1$. Hence, all the posterior probabilities form a $k \times m$ decision matrix as follows:

\begin{equation}
P_{j} =
\begin{bmatrix}
p^{1}_{1j} &  p^{2}_{1j} & \cdots & p^{m}_{1j}\\
p^{1}_{2j} &  p^{2}_{2j} & \cdots & p^{m}_{2j}\\
\vdots &  \vdots & \ddots & \vdots\\
p^{1}_{kj} &  p^{2}_{kj} & \cdots & p^{m}_{kj}\\
\end{bmatrix}
\label{decision_matrix}
\end{equation}

Since the importance of different classifiers might be different, we assign a wight $w_{i}$ to the decision vector (i.e., posterior probabilities vector) of each classifier $C_{i}$, where $i \in \{1, 2, \dots, k\}$. Let $P_{m}(j)$ denote the sum of the posterior probabilities, that sample $j$ belonging to class $m$, of all the classifiers. Then, we have

\begin{equation}
P_{m}(j)=\sum_{i=1}^{k} w_{i} \cdot p^{m}_{ij}
\label{final_posterior_probability}
\end{equation}

The final decision (i.e., class) $D(j)$ of sample $j$ is determined by the maximum posterior probabilities sum:

\begin{equation}
D(j) = \textit{$\underset{i}{max} \ P_{i}(j)$}, \ i \in \{1, 2, \dots, m\}
\label{final_decision}
\end{equation}

In our networked AI, we adopt the average fusion strategy in the multi-classifier fusion, where all the classifiers use the same static weight as $\frac{1}{k}$.

\subsection{Split Artificial Intelligence Decision Unit} \label{sec:methodology_du}
The core component of our proposed SAIA framework is the decision unit component, which controls whether a client-side captured sample (e.g., image) would be sent to the networked AI, or would be processed by the embedded AI. We adopt a meta-information based algorithm selection approach in the design of our decision unit component. In the training phase, we (i) use a set of meta-information generation samples (apart from the training/testing samples of the embedded AI and the networked AI) to generate a set of meta-information (e.g., various features directly extracted from each sample, and the soft predicted probabilities by the embedded AI for each sample), (ii) use our customized decision rule to generate the true label (i.e., ``kept for the embedded AI'' or ``sent to the networked AI'') of each sample, and (iii) use the meta-information and true labels of those samples to train a lightweight binary classifier as the decision unit. In the testing phase, our framework extracts the same set of meta-information from each testing sample, and tests it through the pre-trained decision unit to determine whether sending the sample to the server or not.

To be simplified, we use the soft predicted probabilities provided by the embedded AI as the meta-information, and use gradient boosted trees to build the decision unit classifier. In this work, we adopt a basic decision rule: (i) if communication resources are available, we will send the meta-information of given sample to the decision unit, and then, if the embedded AI and the networked AI produce different predicted results (e.g., classes of a healthcare application) and the networked AI is correct, the sample will be sent to the server (i.e., using the networked AI), otherwise, it will be kept on the client (i.e., using the embedded AI); (ii) if the communication resources are not available, we will keep everything on the client (i.e., using the embedded AI).

Other than just using a ``yes or no'' binary decision, we also design the decision unit to utilize a parameter, namely $\epsilon$, to tune the criteria of how much data could be sent to the networked AI. To be specific, we propose a weighted loss function, where the gradients of the samples that should be ``sent to the networked AI'' are scaled by a parameter $\epsilon$. In the binary classification problem of the decision unit, let us denote the samples that should be ``sent to the networked AI'' as the positive class, i.e., $(y_i = 1)$, and denote the samples that should be ``kept for the embedded AI'' as the negative class, i.e., $(y_i = 0)$.

Then, the objective function of the gradient boosted trees at iteration $t$ is optimized by the simplified second-order approximation \cite{chen2016xgboost} of the original loss function, which is defined as below:

\begin{equation}
\begin{split}
L^{(t)} \approx& \sum_{i=1}^{n} S(y_i) \cdot \bigg[l(y_i, \hat{y}^{(t-1)}) \cdot  g_i \cdot f_t(x_{i}) \\ &+ \frac{1}{2} \cdot h_i \cdot f_{t}^2(x_i)
\bigg] + \Omega (f_t)
\end{split}
\label{GBT_Loss}
\end{equation}

where $l(y_i, \hat{y}^{(t-1)})$ is the cross-entropy loss function, $g_i = \partial_{\hat{y}^{(t-1)}} l(y_i, \hat{y}^{(t-1)})$ and $h_i = \partial^2_{\hat{y}^{(t-1)}} l(y_i, \hat{y}^{(t-1)})$ are the gradient and hessian statistic of the loss function and $\Omega(f_t)$ is the penalized term. Our customized function $S(y_i)$ is defined as below:

\begin{equation}
S(y_i) = \begin{cases} \epsilon, & \mbox{if } y_i = 1 \\ 1, & \mbox{if } y_i = 0 \end{cases}
\label{GBT_epsilon}
\end{equation}
where $\epsilon$ is a predefined hyperparameter that could be adjusted to increase/decrease the expected true positive rate of the decision unit, so that to optimize the amount of data that should be sent to the server.

\section{Experimental Evaluation} \label{sec:experimentalEvaluation}
\subsection{Experiment Environment}
We implemented our embedded AI on a Google Pixel 4 XL smartphone that has a Qualcomm Snapdragon 855 chip-set, 6GB RAM and Android 10.0 OS.
Our network-AI was implemented and performed on a server with Intel$\circledR$Core$^{TM}$ i9-7980XE@2.60GHz CPU, 128GB RAM and 4 GTX 1080Ti 11GB GPUs.

\subsection{Experiment Datasets} \label{sec:experimentalEvaluation_data}
We investigated two popular benchmark healthcare image datasets in our experimental evaluation: (i) \textbf{I}nternational \textbf{S}kin \textbf{I}maging \textbf{C}ollaboration Challenge 2019 (ISIC 2019) \cite{tschandl2018ham10000, codella2018skin, combalia2019bcn20000} and (ii) Onychomycosis dataset \cite{han2018deep}. ISIC 2019 has training and testing sets with overall 33,569 images. Since the ground truth of the testing data was not available, we only employed its original training data in our evaluation. It contains 25,331 images of 8 skin lesion diseases (i.e., 8 classes): melanoma (4,522), melanocytic nevus (12,875), basal cell carcinoma (3,323), actinic keratosis (876), benign keratosis (2,624) dermatofibroma (239), vascular lesion (253) and squamous cell carcinoma (628). We randomly split 80\%, 5\% and 15\% as training data, meta-information data and testing data, respectively. In order to have enough data to train the decision unit and base classifiers, data augmentation was applied to enlarge the training data and meta-information data by performing different rotation degrees (i.e., 90, 180 and 270), horizontal flipping and combinations of both. Thus, the training data and meta-information data became 81,020 and 10,400 in total, respectively. Regarding onychomycosis dataset, which contains 53,794 region-of-interest extracted abnormal (34,014) and normal (19,780) fingernail images. We split it into training/meta-information/testing by the ratio of 70\%/10\%/20\%, respectively. Due to enough number of samples, we only performed horizontal flipping on the meta-information dataset. Note that our embedded AI and networked AI were trained on both original and augmented images.

\subsection{Embedded/Networked AI and Decision Unit Preparation} \label{experimentalPreparation}
{\bf Embedded AI.} In the preparation of the Embedded AI, for each dataset, we trained three conventional machine learning classifiers (i.e., SVM, RF and DART). To figure out the optimal set of hyperparameters for each classifier, we performed 5-fold cross validation for each classifiers. Table~\ref{tab_mls} shows the performance (i.e., accuracy) of three classifiers on two datasets, where with the optimal settings, DART classifier outperforms SVM and RF classifiers on both datasets. Furthermore, while applying One-vs-All strategy for training SVM and RF classifiers in skin lesion dataset that has 8 classes, using SVM and RF result in considerably much larger model size than using DART, that consumes more storage space of the mobile and IoT devices. Also, since DART uses softmax function as the objective function, DART classifier directly provides the soft predicted probabilities of each sample, which would be taken as the meta-information to train our decision unit (as described in Section~\ref{sec:methodology_du}). Therefore, we decided to deploy the DART classifier as the embedded AI for both datasets in the rest of our experiments.

\begin{table}
  \caption{The performance (accuracy in \%) of conventional machine learning classifiers (i.e., the embedded AI) on Skin Lesion and Onychomycosis datasets.}
  \label{tab_mls}
  \scalebox{0.92}{
  \begin{tabular}{ccc}
    \toprule
    \textbf{Embedded-AI Models} & \textbf{Skin Lesion} & \textbf{Onychomycosis} \\
    \midrule
    DART \cite{rashmi2015dart} & 75.89 & 78.61 \\
    \midrule
    SVM \cite{suykens1999least} & 68.90 & 72.77 \\
    \midrule
    RF \cite{liaw2002classification} & 65.04 & 70.07  \\

    \bottomrule
    \end{tabular}}
\end{table}

{\bf Networked AI} We evaluated twelve different CNN architectures (as shown in Table \ref{tab_cnn}) on the server side with pre-trained weights on ImageNet \cite{imagenet_cvpr09}. Different networks expect different input sizes: 331$\times$331 for PNASNet-5-Large and NASNet-A-Large; 320$\times$320 for ResNeXt101-32$\times$16d; 299$\times$299 for InceptionResNet-V2, Xception, Inception-V4 and Inception-V3; 224$\times$224 for SENet154, SE-ResneXt101-32, EfficientNet-B7, Dual Path Net-107$\times$4d and ResNet152. All the networks were fine-tuned in Pytorch, using SGD optimizer with learning rate 0.001 (degraded after 20 epochs by 0.1) and momentum 0.9. We stopped the training process either in 40 epochs or the validation accuracy failed to improve for over 7 consecutive epochs. To keep the same batch size 32 in each evaluation, and due to the memory constraint of single GPU, certain networks were trained parallelly with multiple GPUs: PNASNet-5-Large (4), NASNet-A-Large (4), ResNext101-32$\times$16d (4), SENet154 (2), EfficientNet-B7 (2) and Dual Path Net-107 (2). The performance result of each base CNN classifier on each dataset has been shown in Table \ref{tab_cnn}. As described in Section~\ref{sec:methodology_networked}, we utilize the multi-classifier fusion of those twelve advanced CNN architectures as our networked AI aiming to provide the SOTA performance of each dataset/application.

\begin{table}
  \caption{The performance (accuracy in \%) of the base classifiers of 12 CNN architectures on Skin Lesion and Onychomycosis datasets.}
  \label{tab_cnn}
  \scalebox{0.85}{
  \begin{tabular}{ccc}
    \toprule
    \textbf{Networked-AI Models} & \textbf{Skin Lesion} & \textbf{Onychomycosis} \\
    \midrule
    SENet154 \cite{hu2018squeeze} & 88.00 & 92.06 \\
    \midrule
    PNASNet-5-Large \cite{liu2018progressive} & 87.87 & 92.36 \\
    \midrule
    NASNet-A-Large \cite{zoph2016neural} & 87.79 & 91.82  \\
    \midrule
    ResNeXt101-32$\times$16d \cite{xie2017aggregated} & 87.76 & 91.69 \\
    \midrule
    SE-ResneXt101-32$\times$4d \cite{hu2018squeeze} & 87.55 & 91.99\\
    \midrule
    InceptionResNet-V2 \cite{szegedy2017inception} & 87.53 & 91.59\\
    \midrule
    Xception \cite{chollet2017xception} & 87.18 & 91.65 \\
    \midrule
    EfficientNet-B7 \cite{tan2019efficientnet} & 86.78 & 92.36\\
    \midrule
    Dual Path Net-107 \cite{chen2017dual} & 86.23 & 91.79 \\
    \midrule
    Inception-V4 \cite{szegedy2017inception} & 85.99 & 92.02 \\
    \midrule
    Inception-V3 \cite{szegedy2016rethinking} & 85.41 & 91.68 \\
    \midrule
    ResNet152 \cite{he2016deep} & 84.00 & 92.28 \\
    \bottomrule
    \end{tabular}}
\end{table}

{\bf Decision Unit} As presented in Section~\ref{sec:methodology_du}, we prepared the decision unit for both datasets accordingly. To evaluate the effectiveness of the tuning hyperparameter $\epsilon$, we use a discrete set of integers ranging from $0$ (i.e., meaning no decision unit deployed) to $100$ for $\epsilon$.

\begin{figure*}[!ht]
  \centering
     \subfloat[]{\includegraphics[width=0.33\textwidth]{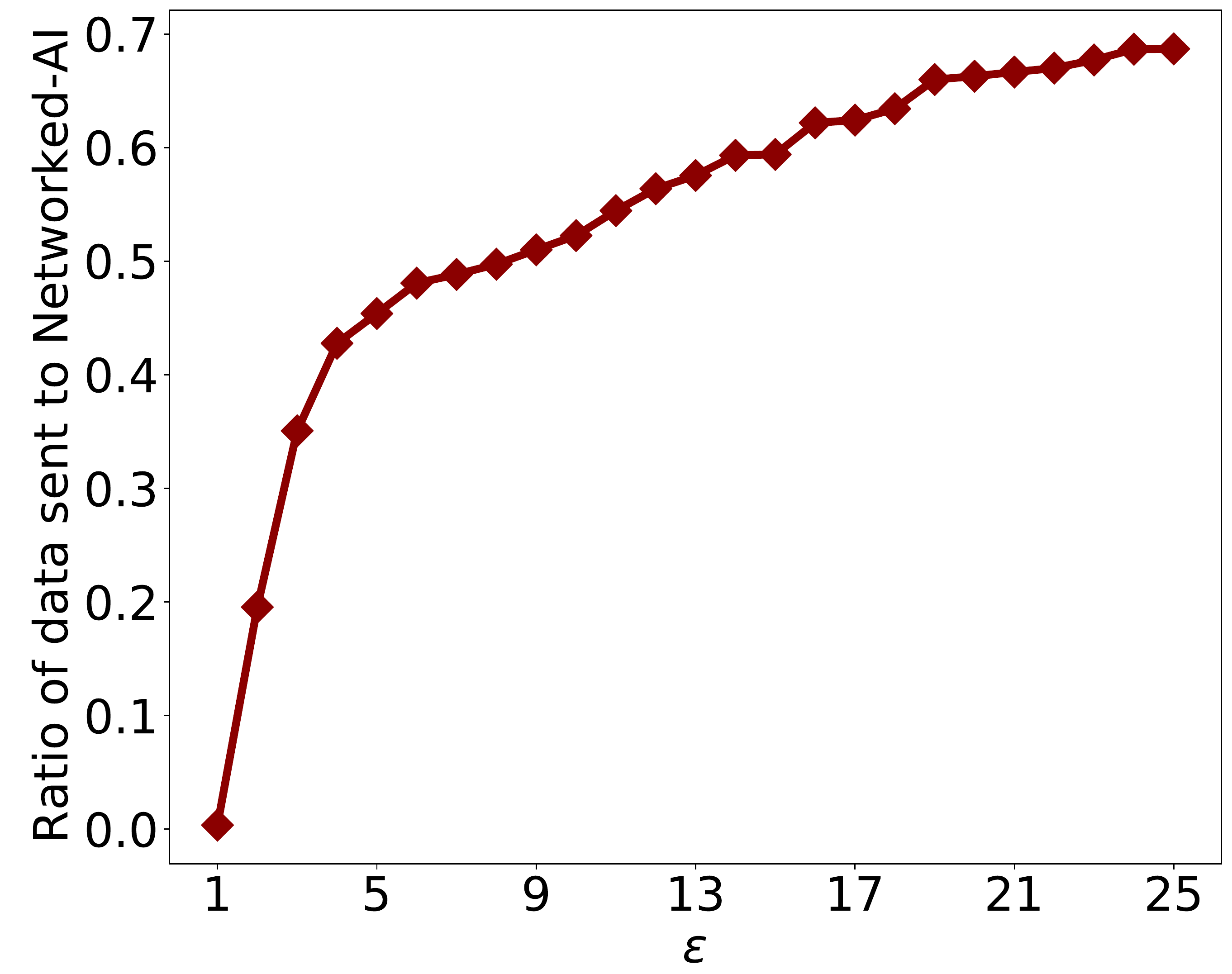} \label{fig:saia_skin_a}}
     \subfloat[]{\includegraphics[width=0.33\textwidth]{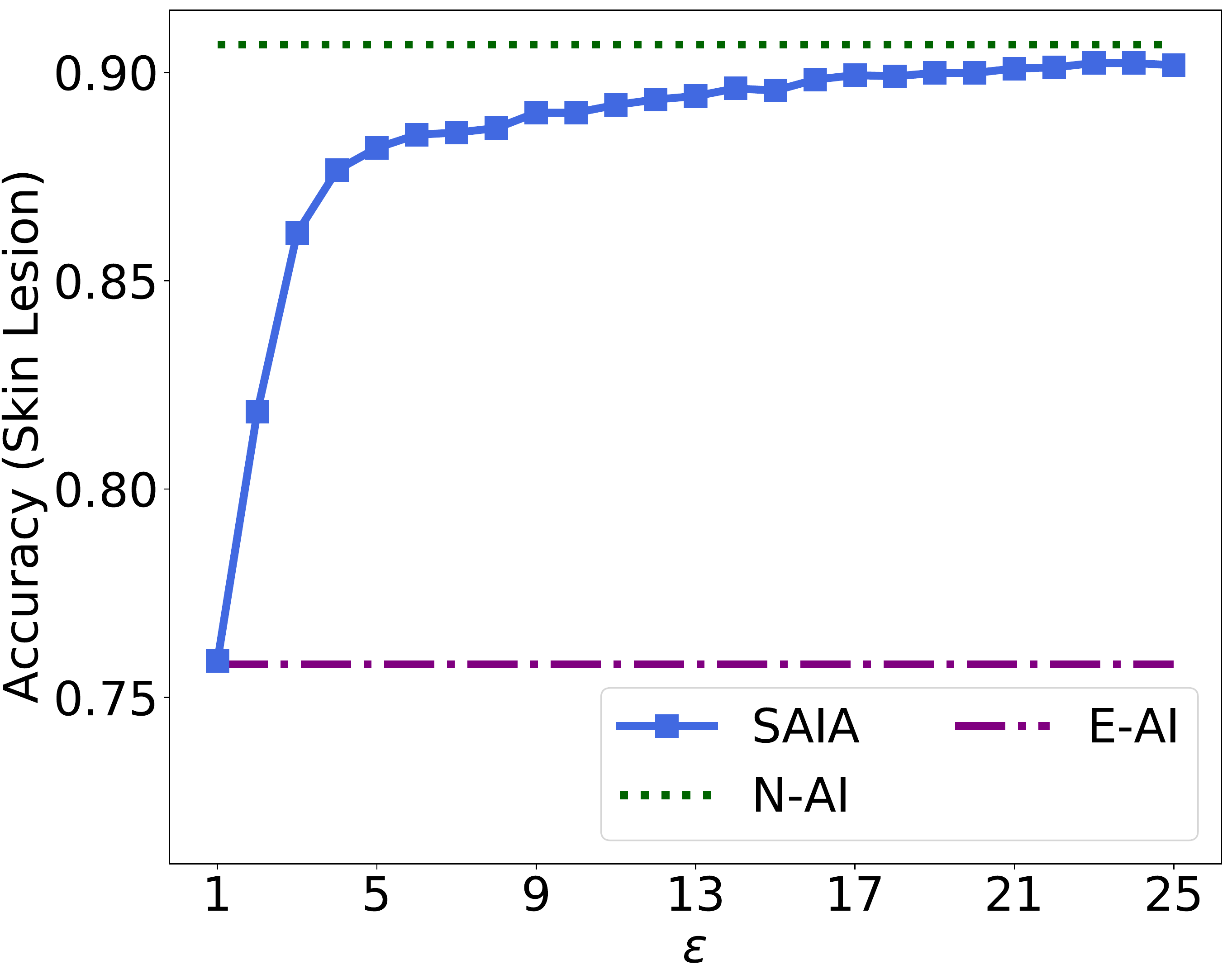} \label{fig:saia_skin_b}}
     \subfloat[]{\includegraphics[width=0.33\textwidth]{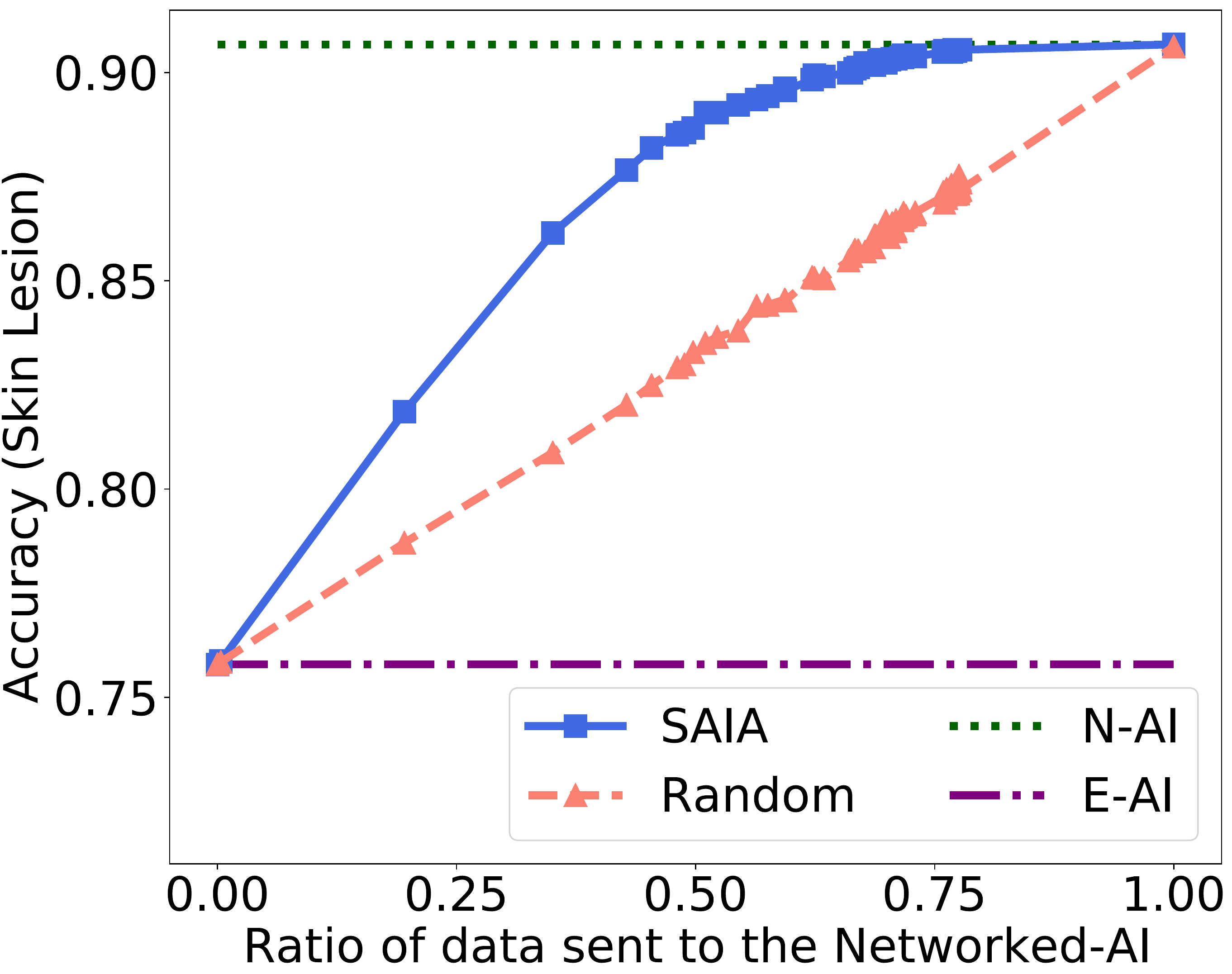} \label{fig:saia_skin_c}}
      \caption{Effectiveness Analysis of Skin Lesion Classification}
  \label{fig:saia_skin}
\end{figure*}

\begin{figure*}[!ht]
  \centering
     \subfloat[]{\includegraphics[width=0.33\textwidth]{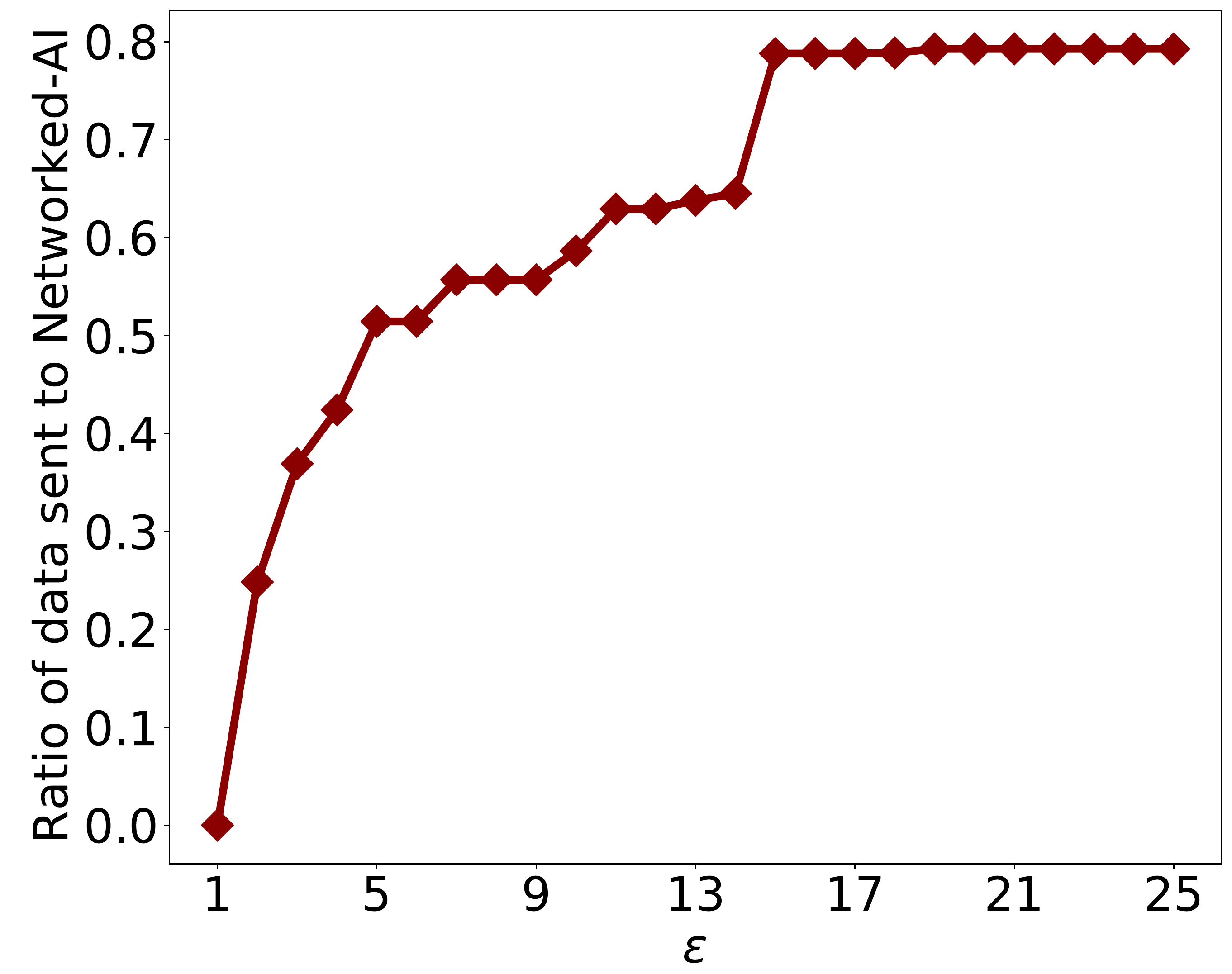} \label{fig:saia_fungus_a}}
     \subfloat[]{\includegraphics[width=0.33\textwidth]{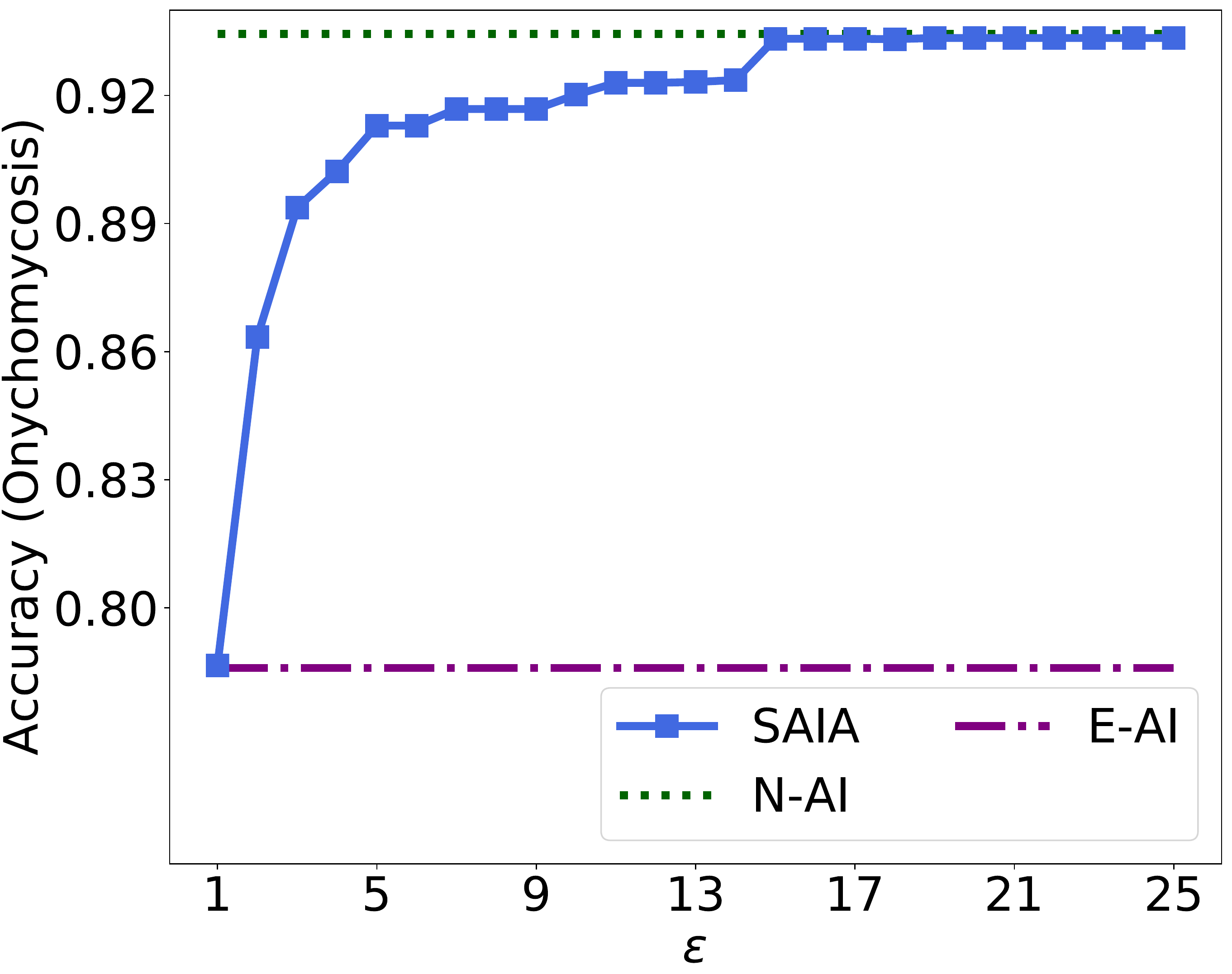} \label{fig:saia_fungus_b}}
     \subfloat[]{\includegraphics[width=0.33\textwidth]{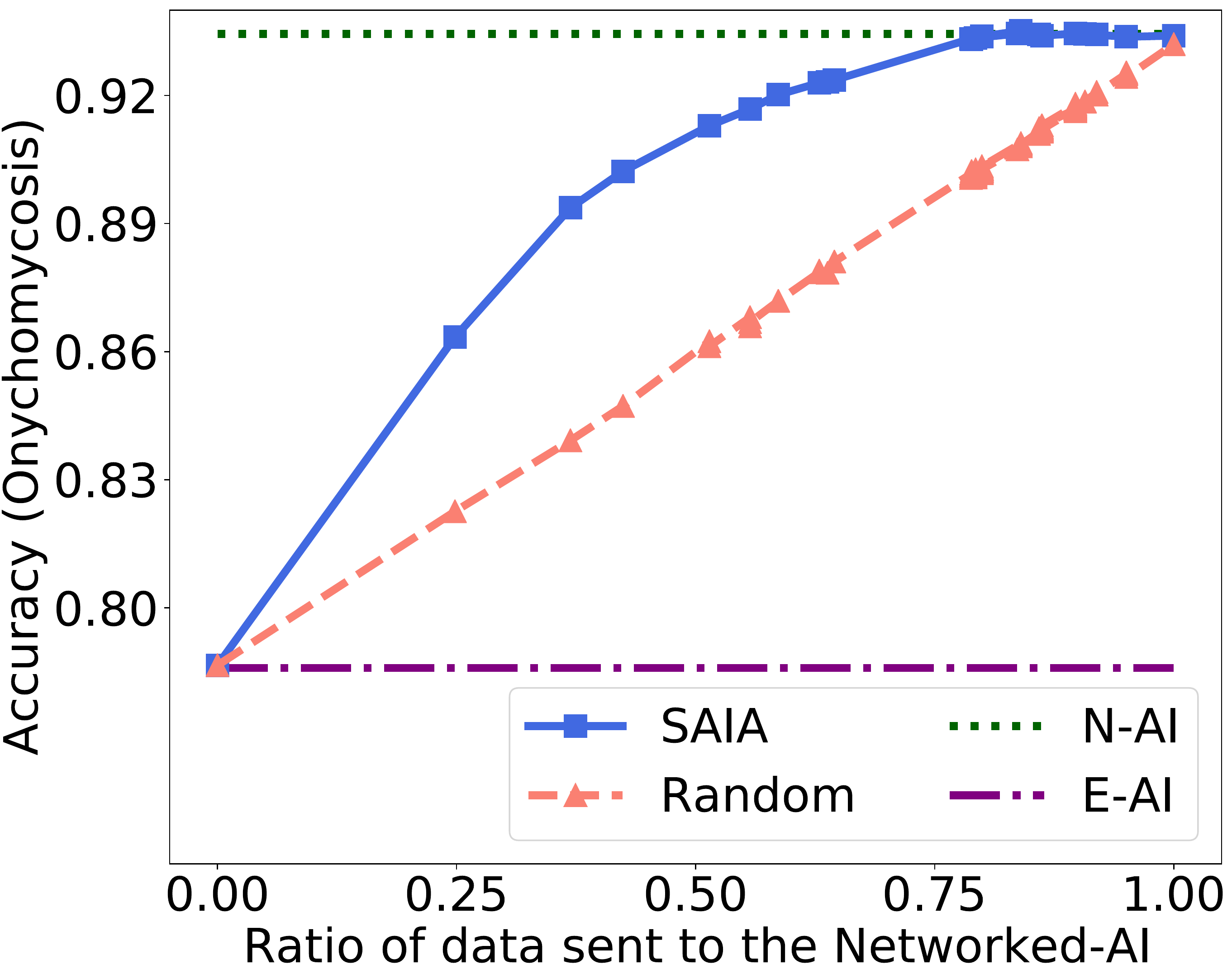} \label{fig:saia_fungus_c}}
      \caption{Effectiveness Analysis of Onychomycosis Classification}
  \label{fig:saia_fungus}
\end{figure*}

\subsection{Effectiveness Analysis}
In this section, we compare the effectiveness of our proposed SAIA framework with three baselines: only using the embedded AI, only using the networked AI, and SAIA but with a randomized decision unit, that randomly determines whether sending a given sample to the server-side or not. All the experiments that utilized the randomized decision unit are performed 100 times and evaluated using the averaged results. Fig.~\ref{fig:saia_skin_a} and Fig.~\ref{fig:saia_fungus_a} show that as we increase the value of $\epsilon$, more and more data would be sent to the networked AI. Also, the curves quickly converges as the $\epsilon$ increasing. Thus, one can tune $\epsilon$ based on the communication resource available to adjust how much data to be sent to the server-side for processing.

Fig.~\ref{fig:saia_skin_b} and Fig.~\ref{fig:saia_fungus_b} illustrate that as we increase the value of $\epsilon$, the accuracies obtained by SAIA for both datasets are also increasing and quickly converged to the SOTA accuracy achieved by the networked AI. For instance, as shown in Fig.~\ref{fig:saia_skin_b}, while $\epsilon=25$, SAIA achieves nearly the same accuracy as the networked AI (i.e., 90\% vs. 90.6\%), but only sends 70\% of the samples to the networked AI (Fig.~\ref{fig:saia_skin_a}). In Fig.~\ref{fig:saia_fungus_b}, while $\epsilon=17$, SAIA achieves exactly the same accuracy as the networked AI (i.e., 93.2\%), but only sends around 80\% of the samples to the networked AI (Fig.~\ref{fig:saia_fungus_a}).

Fig.~\ref{fig:saia_skin_c} and Fig.~\ref{fig:saia_fungus_c} present the comparison among our proposed SAIA, SAIA with a randomized decision unit, only using the embedded AI and only using the networked AI. We observed that (i) our proposed SAIA consistently outperforms the SAIA with a randomized decision unit (other than while keeping all the data at the embedded AI or the network AI); (ii) as sending more samples to the server, compared with the SAIA with a randomized decision unit, the accuracy of our proposed SAIA framework quickly converges to the accuracy of the networked AI (i.e., while sending 75\% samples of the skin lesion dataset, and while sending 80\% samples of the onychomycosis dataset); (iii) while sending nearly half of the samples to the server, our proposed SAIA obtains the highest accuracy advantage over the SAIA with a randomized decision unit, e.g., as shown in Fig.~\ref{fig:saia_skin_c}, while $\epsilon = 10$, almost over half of the skin lesion samples (51\%) were sent to the server, and the difference of the accuracies is 88.87\% (ours) vs. 83.95\% (randomized), and as shown in Fig.~\ref{fig:saia_fungus_c}, while $\epsilon = 5$, 53\% of the onychomycosis samples were sent to the server, and the difference of the accuracies is 92\% (ours) vs. 86\% (randomized). To summarize, our proposed SAIA could control how much data to be sent to the server based on the environment and accuracy requirement. Our framework could also achieve the same accuracy as processing on the server side, while sending much less data to the server side.

\begin{figure}[h]
  \centering
     \subfloat[$\epsilon = 3$]{\includegraphics[width=0.14\textwidth]{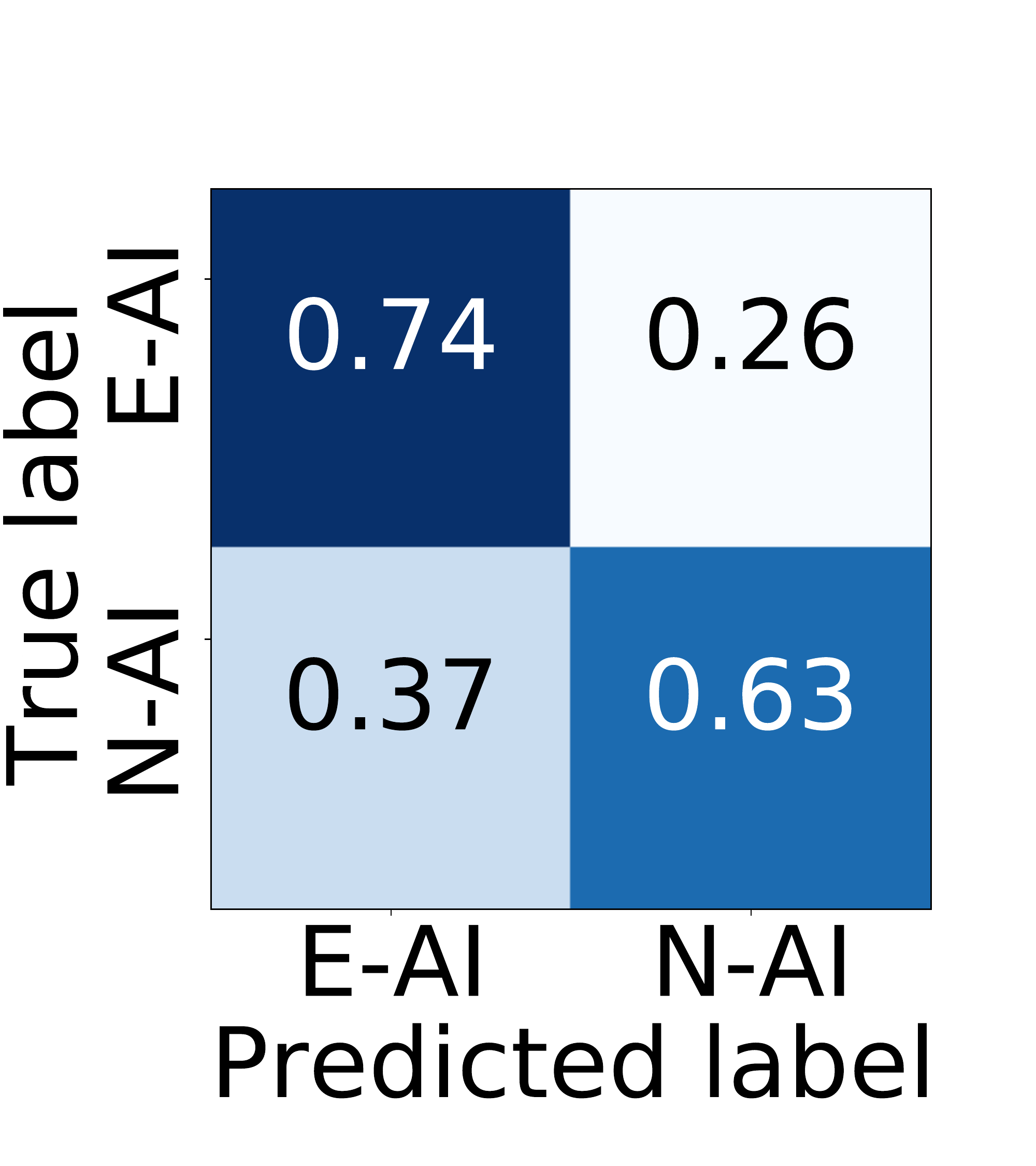}}\hfil
     \subfloat[$\epsilon = 6$]{\includegraphics[width=0.14\textwidth]{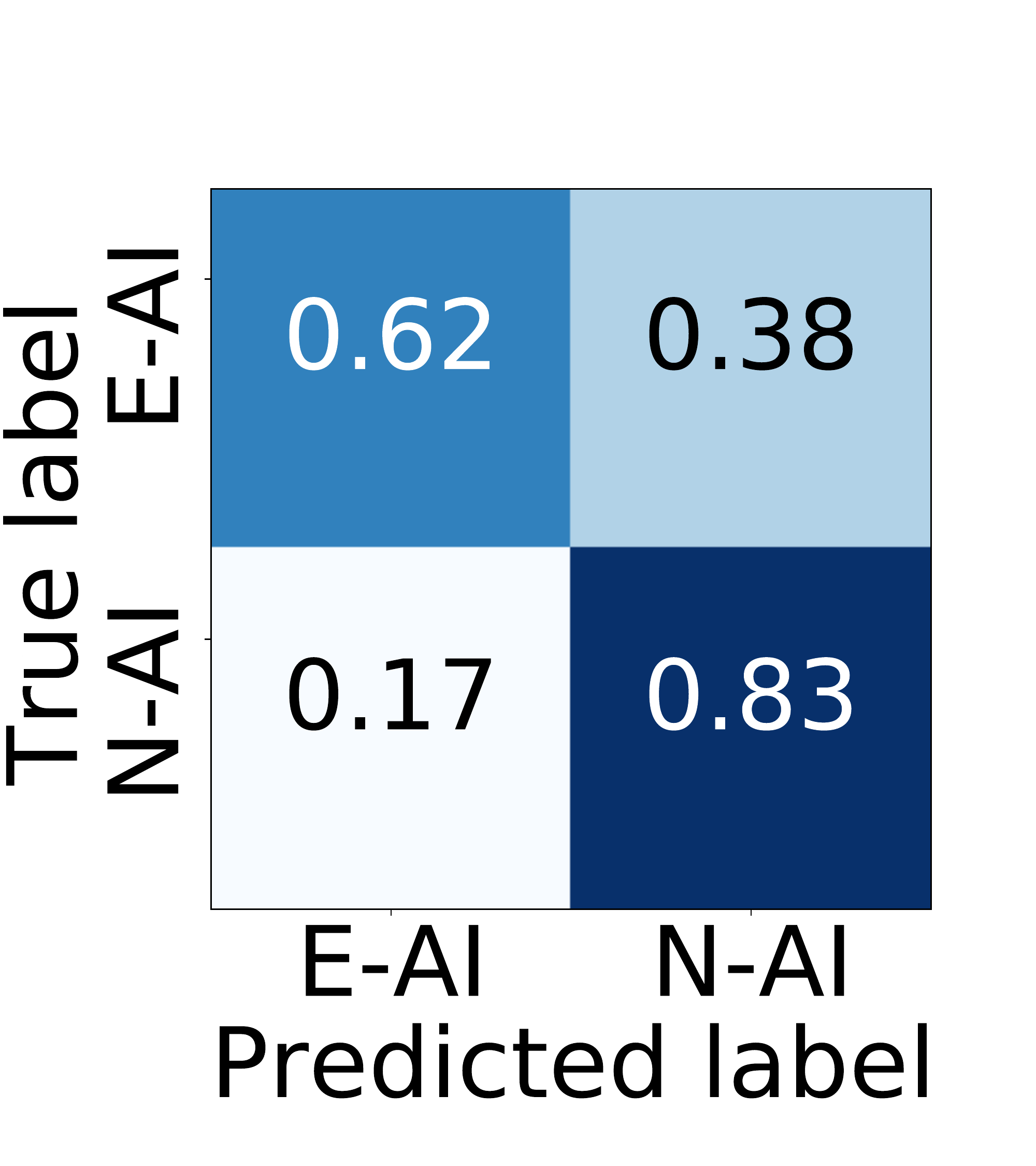}}\hfil
     \subfloat[$\epsilon = 10$]{\includegraphics[width=0.14\textwidth]{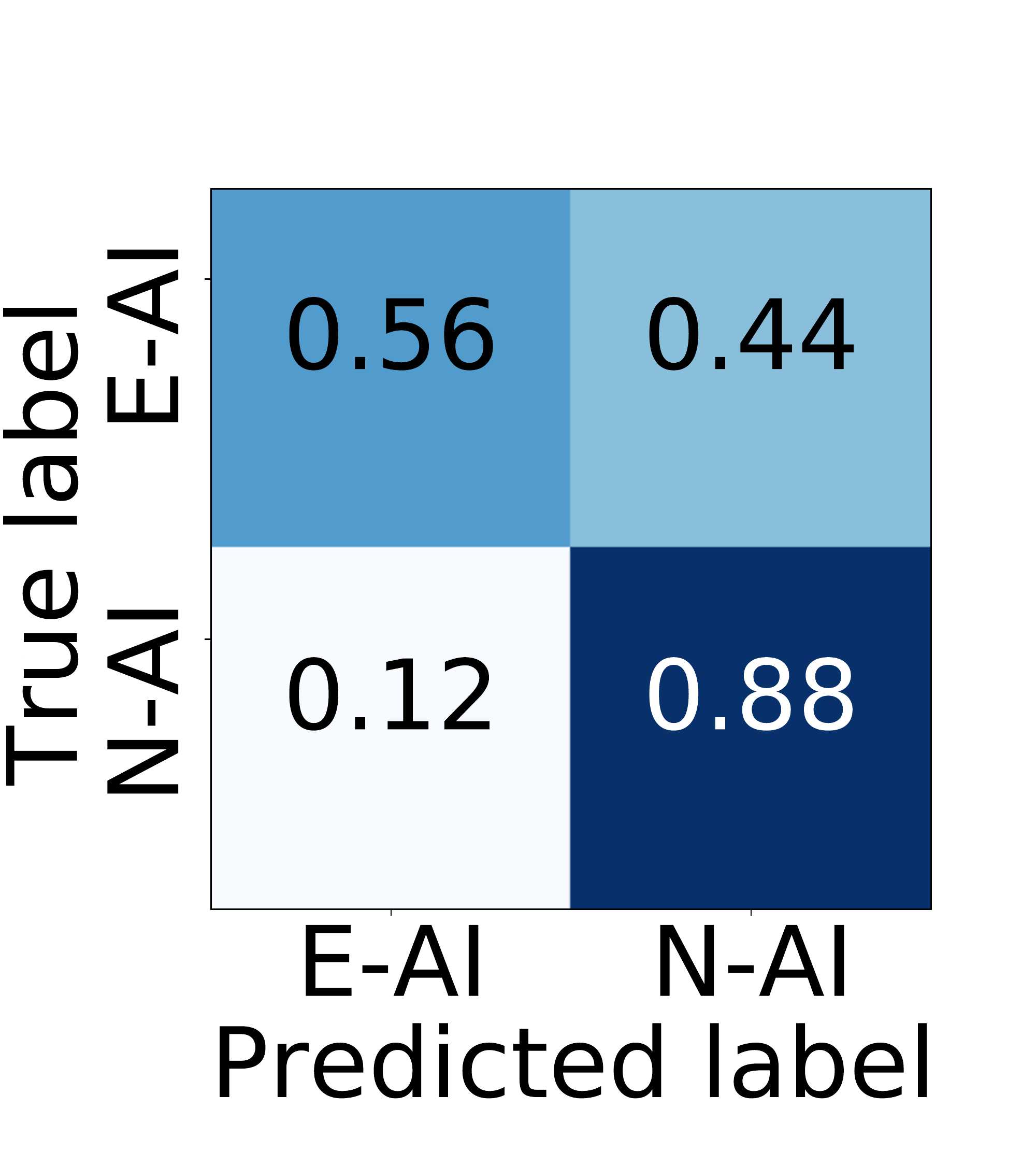}}\hfil
      \caption{Confusion matrix of SAIA decision unit (Skin Lesion)}
  \label{fig:cfm_isic}
\end{figure}

\begin{figure}[h]
  \centering
     \subfloat[$\epsilon = 3$]{\includegraphics[width=0.14\textwidth]{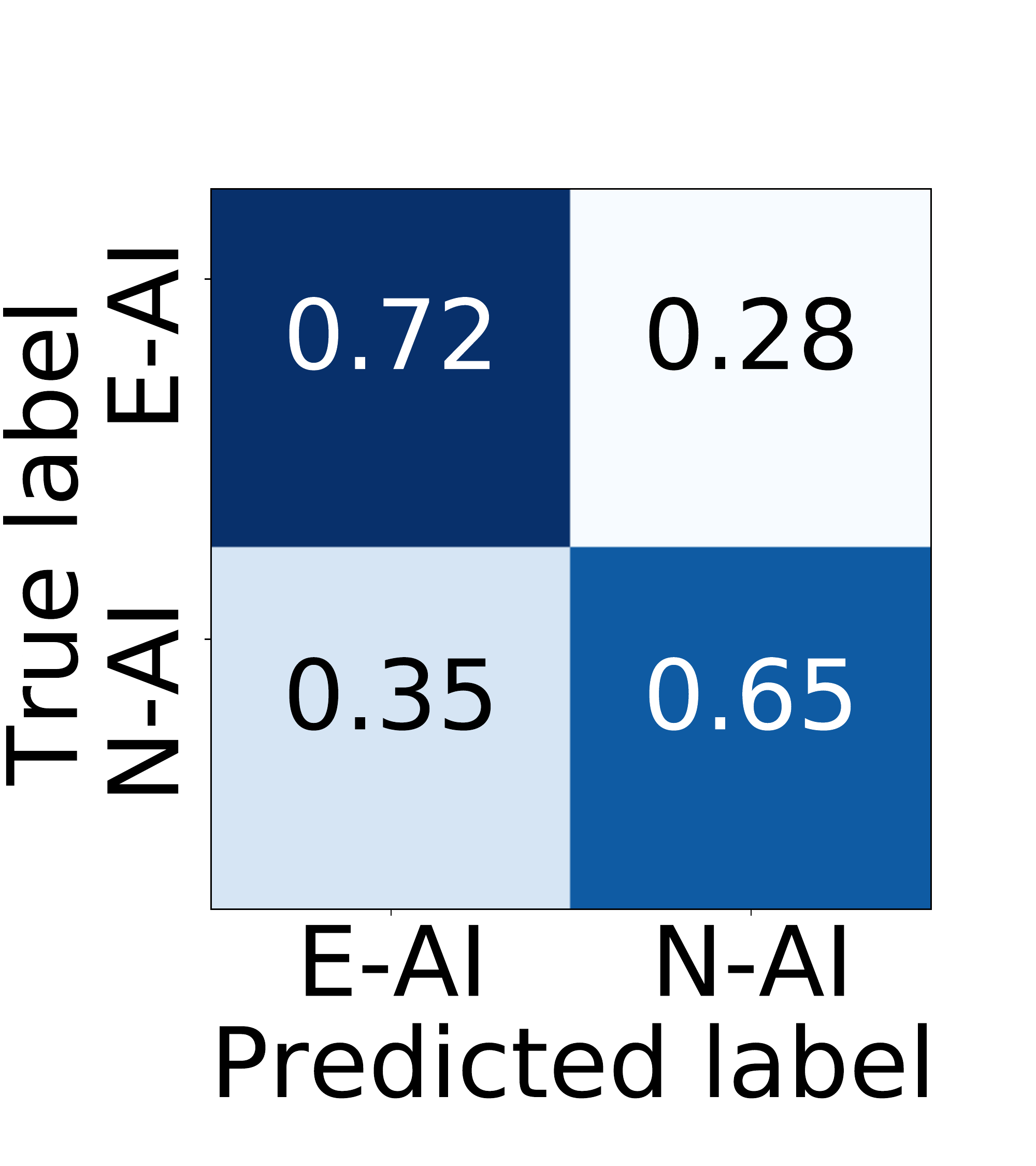}}\hfil
     \subfloat[$\epsilon = 5$]{\includegraphics[width=0.14\textwidth]{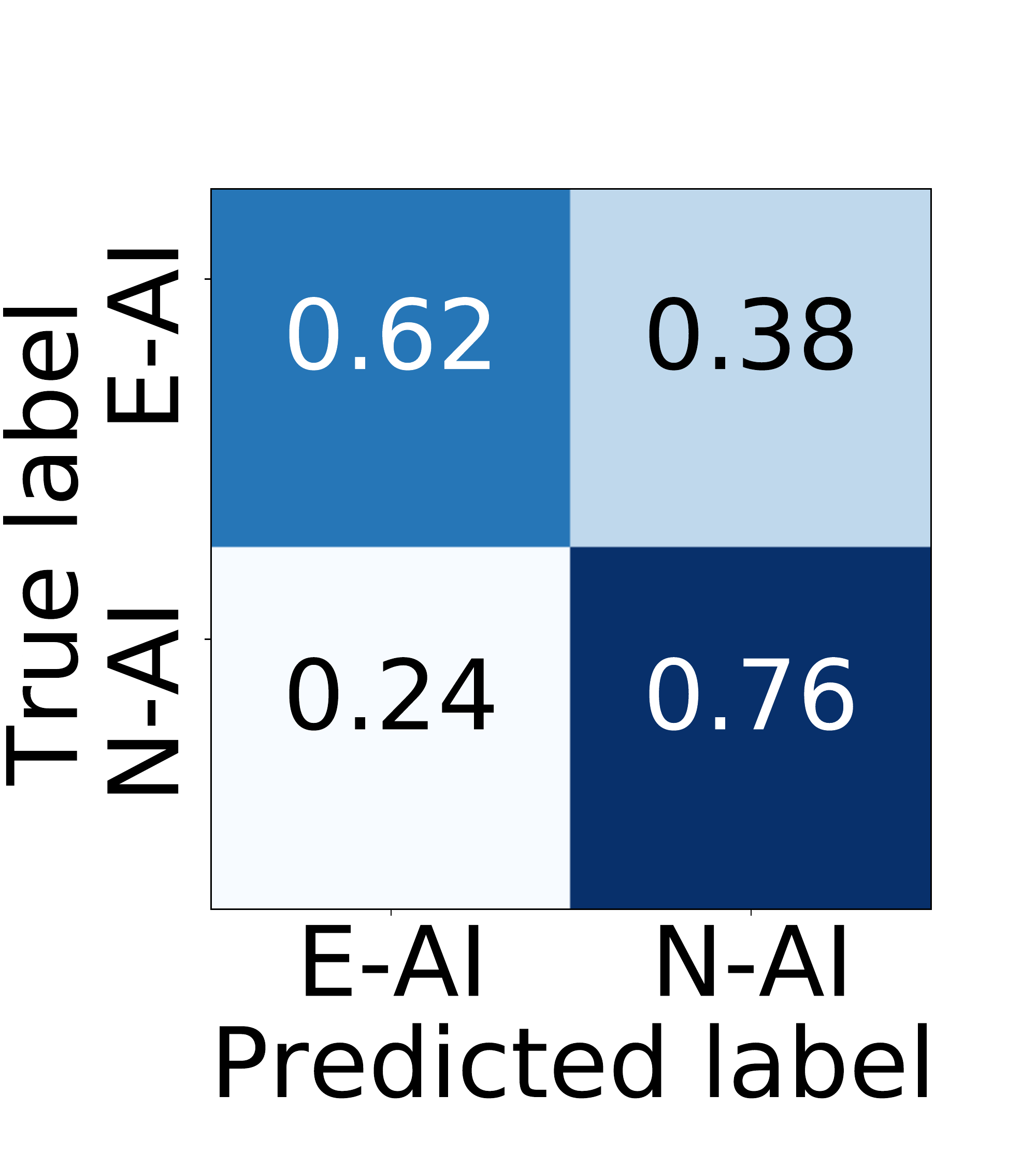}}\hfil
     \subfloat[$\epsilon = 9$]{\includegraphics[width=0.14\textwidth]{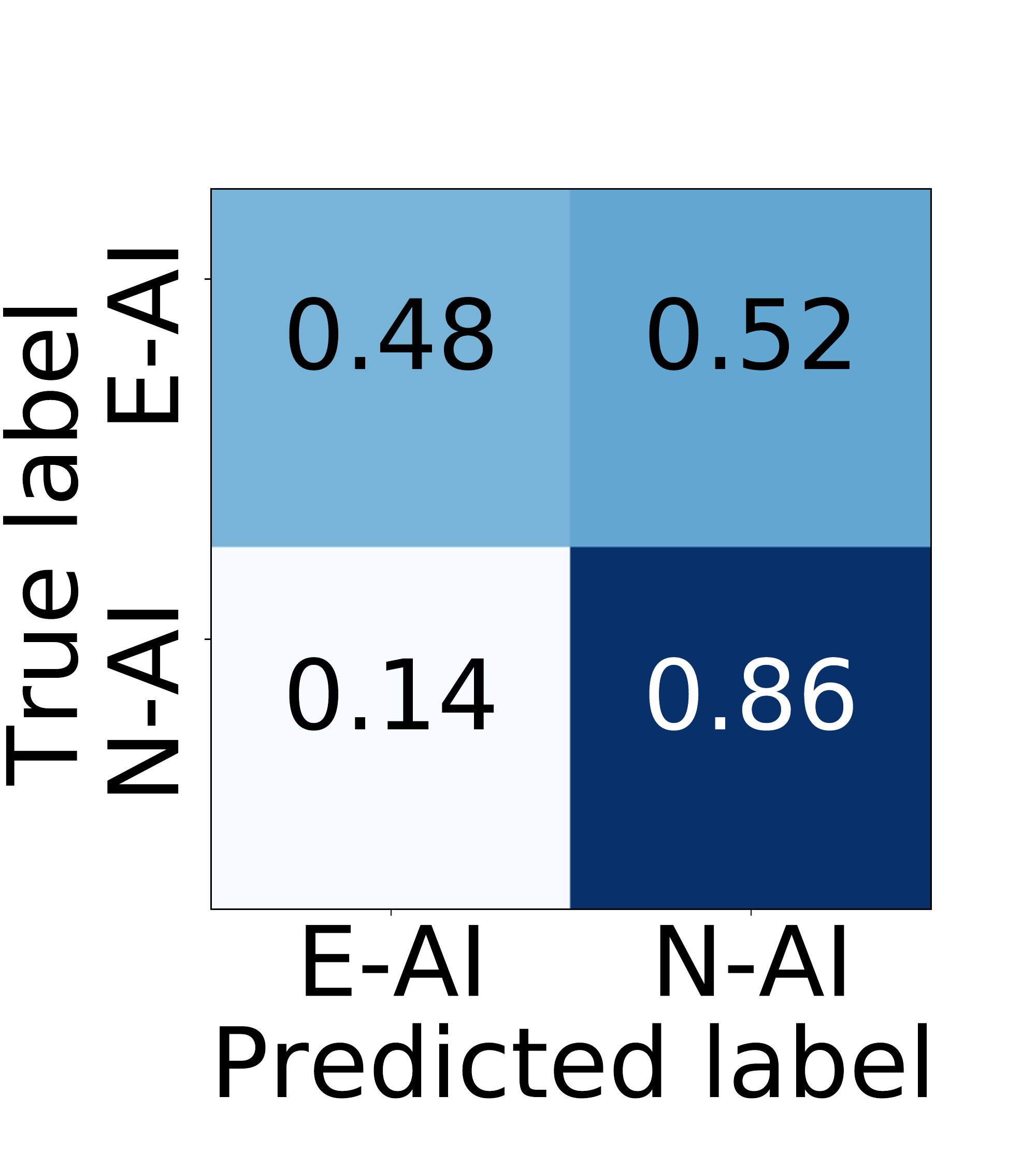}}\hfil
      \caption{Confusion matrix of SAIA decision unit (Onychomycosis)}
  \label{fig:cfm_fungus}
\end{figure}

\begin{figure}[h]
  \centering
    \subfloat[]{\includegraphics[width=0.42\textwidth]{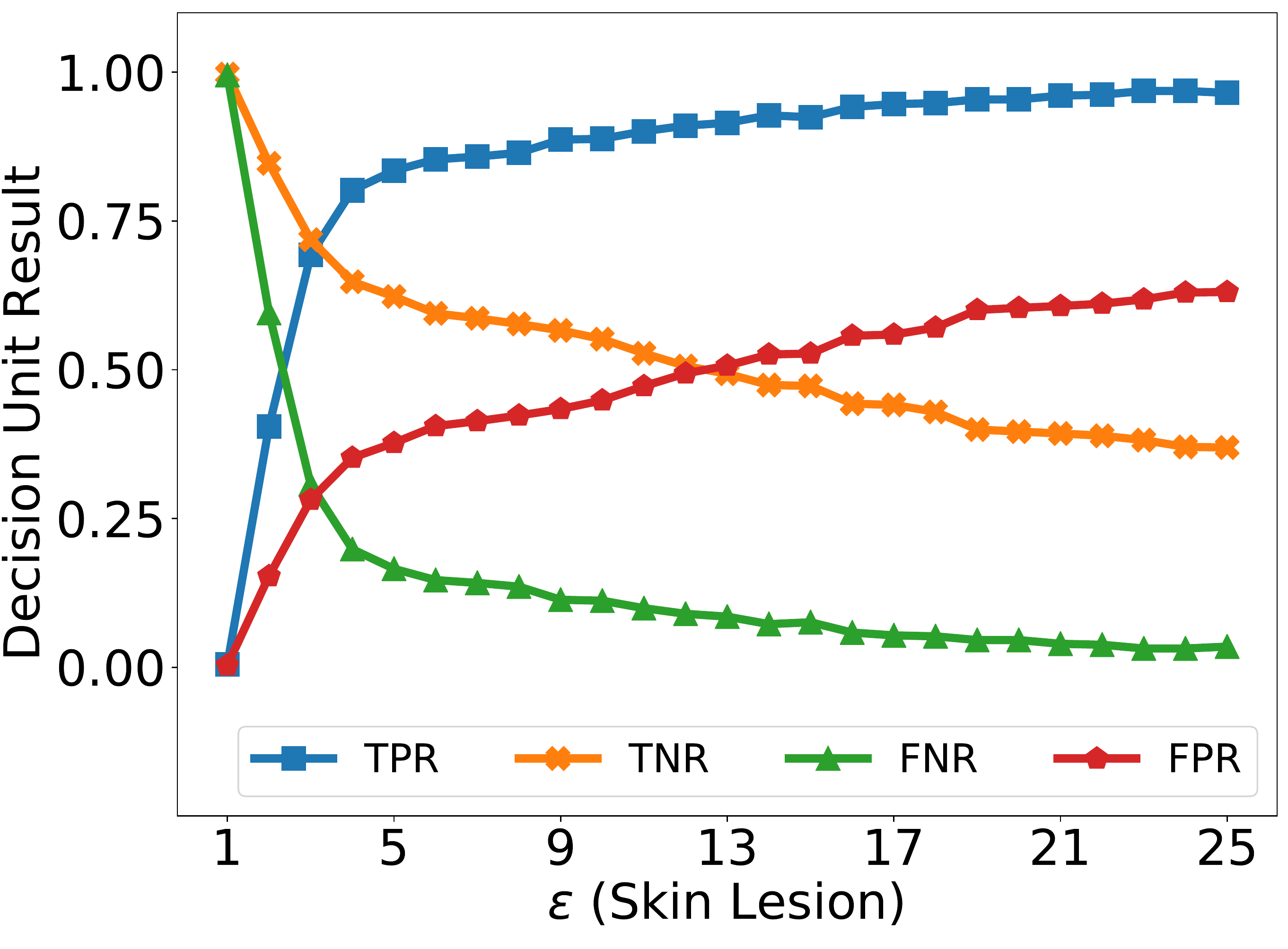}}\hfil
    \subfloat[]{\includegraphics[width=0.42\textwidth]{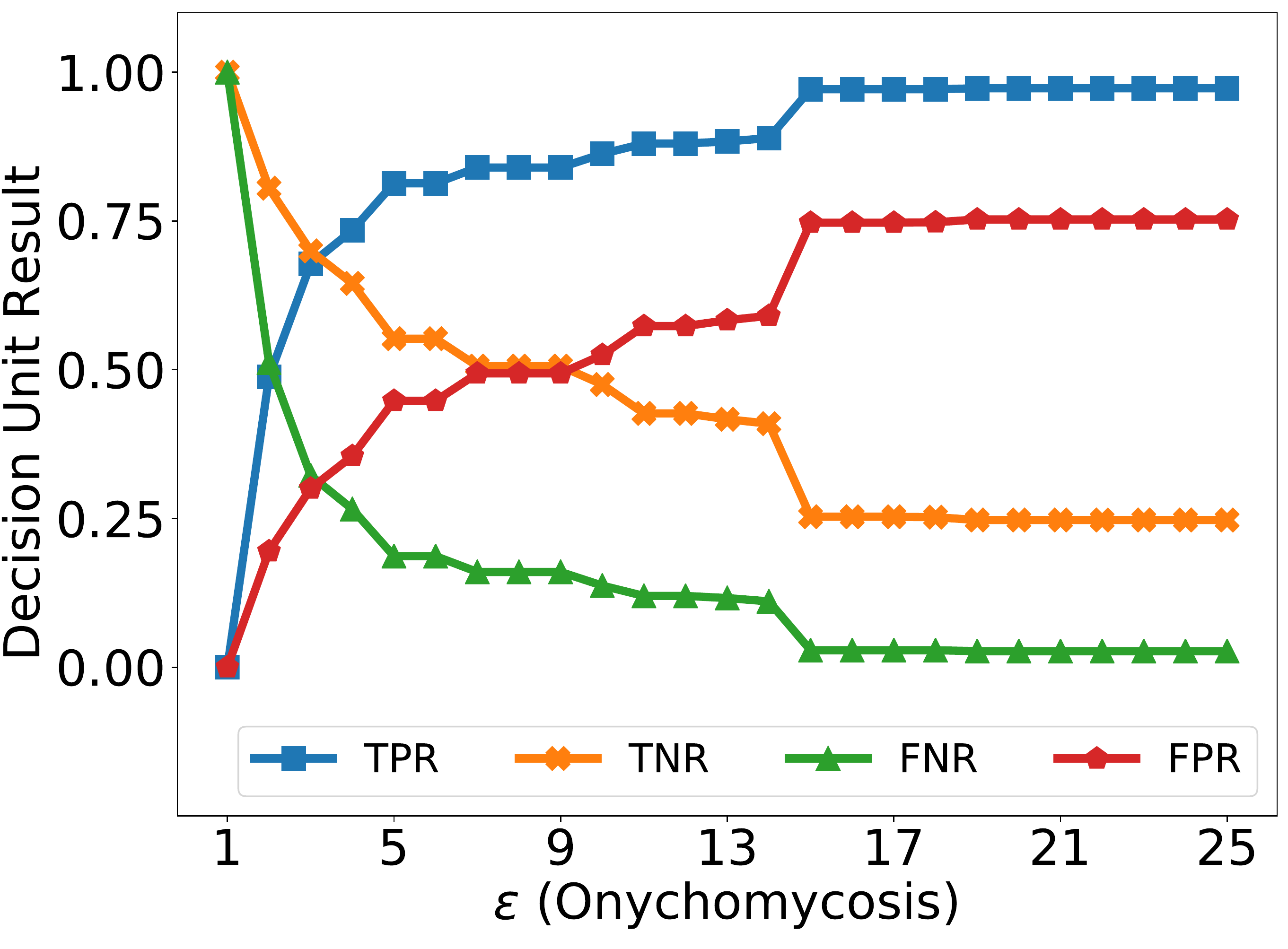}}
    \caption{True Positive Rate (TPR), True Negative Rate (TNR), False Positive Rate (FPR) and False Negative Rate (FNR) of (a) Skin Lesion, (b) Onychomycosis.}
  \label{fig:tptn}
\end{figure}

\subsection{The Performance of Decision Unit and the Effectiveness of the Hyperparameter $\epsilon$}
\begin{figure*}[!t]
  \centering
     \subfloat[]{\includegraphics[width=0.33\textwidth]{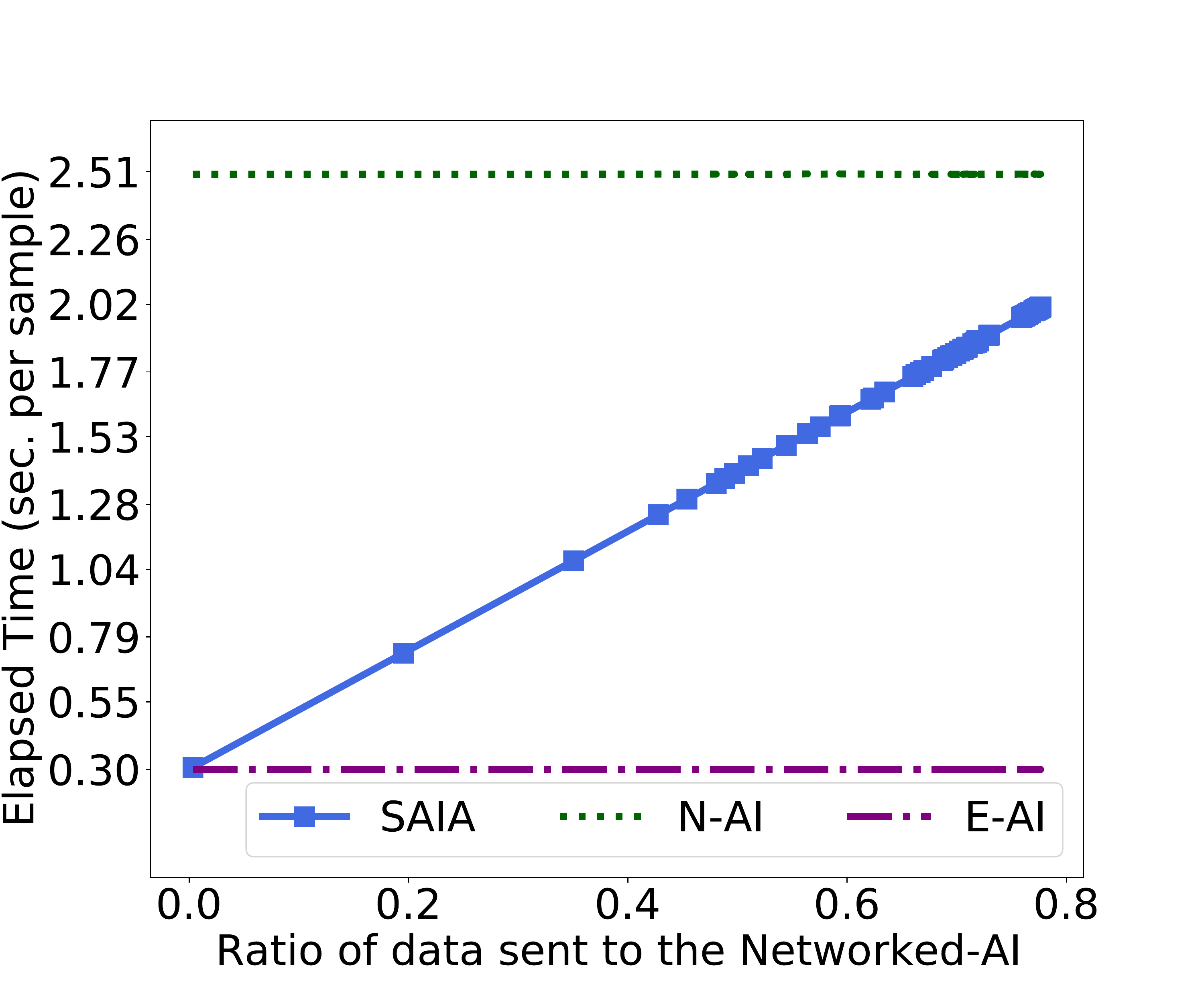} \label{fig:eff_skin_a}}
     \subfloat[]{\includegraphics[width=0.33\textwidth]{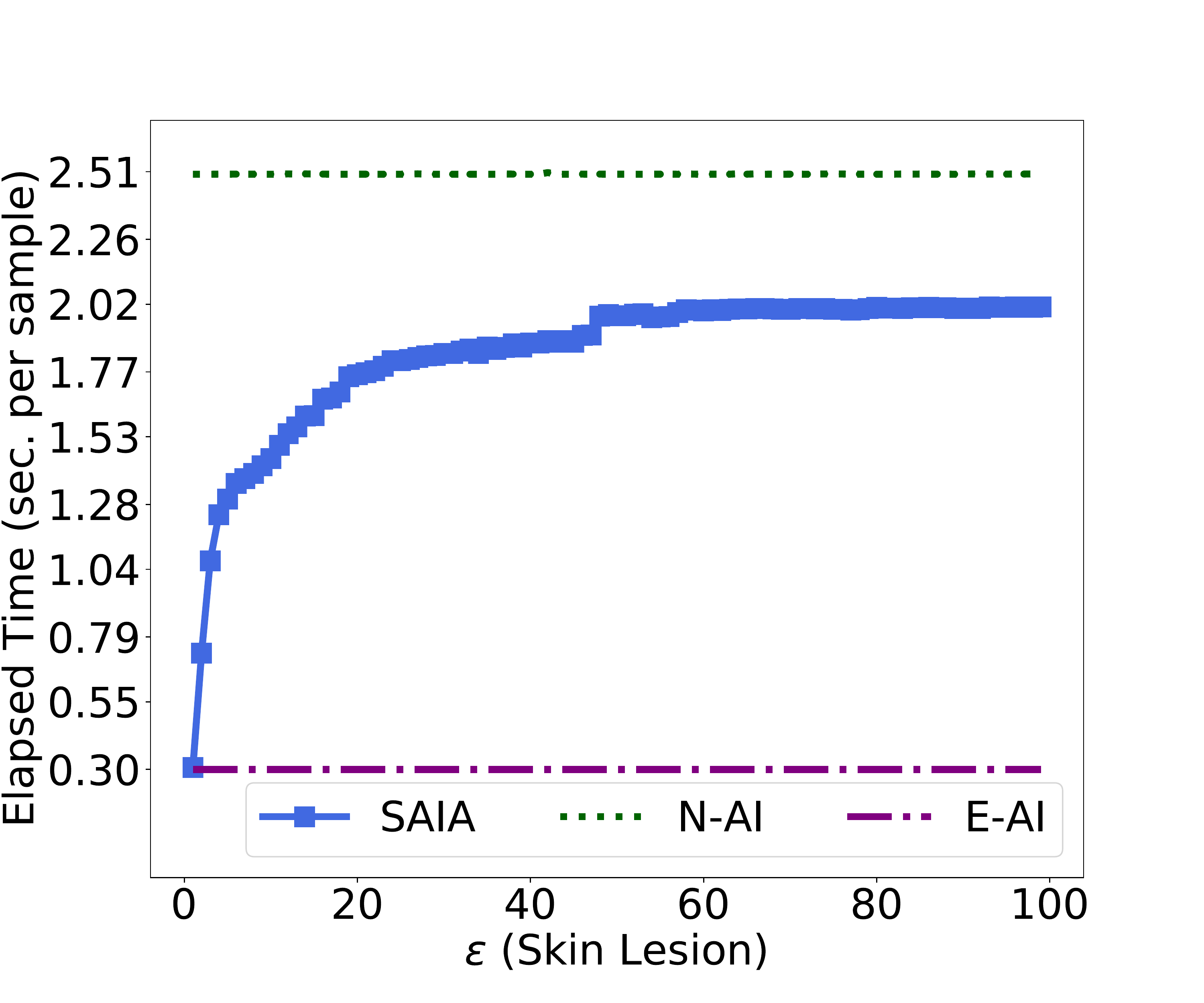}
     \label{fig:eff_skin_b}}
     \subfloat[]{\includegraphics[width=0.33\textwidth]{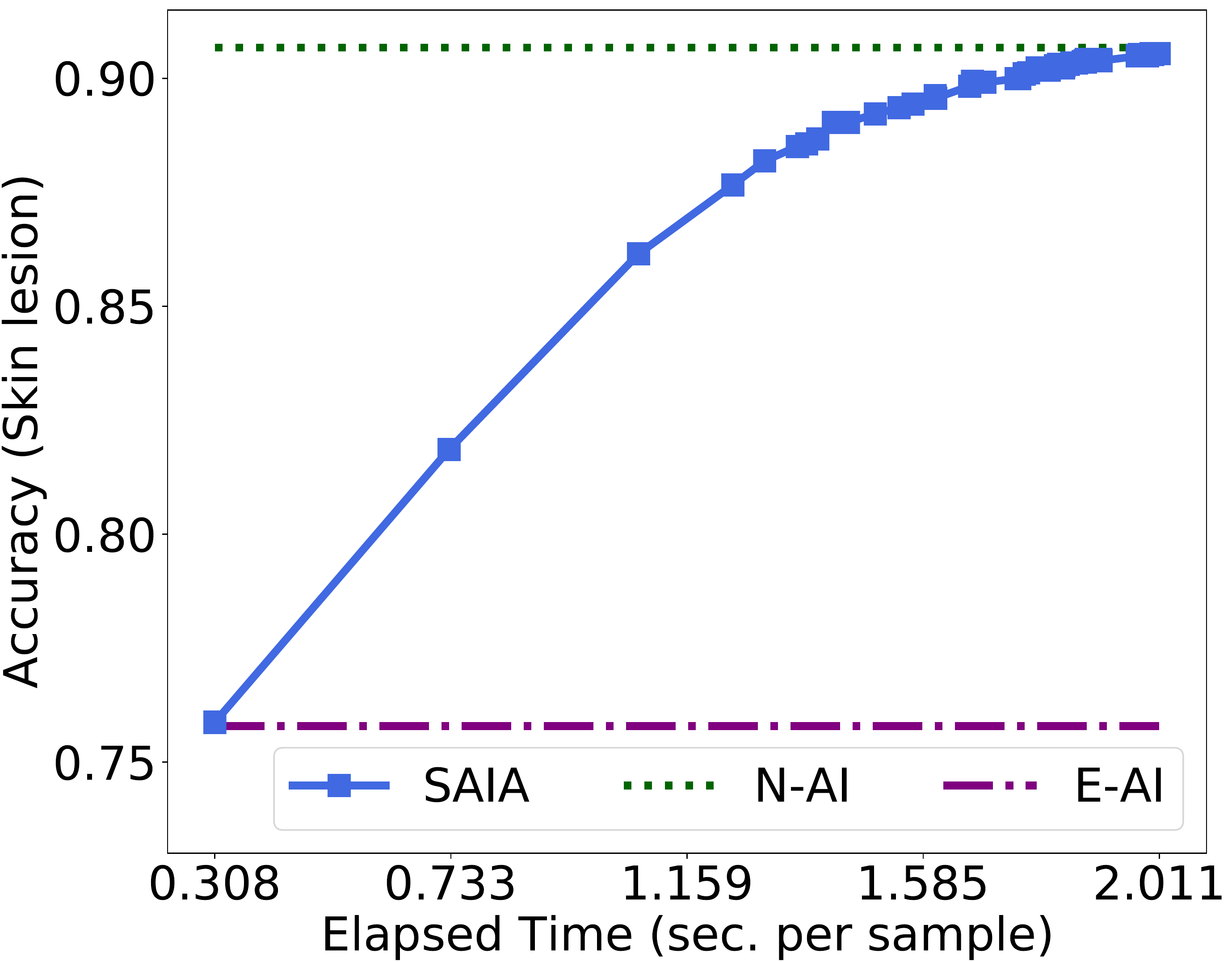}
     \label{fig:eff_skin_c}}
      \caption{Efficiency Analysis Skin Lesion Classification}
  \label{fig:eff_skin}
\end{figure*}

\begin{figure*}[!t]
  \centering
     \subfloat[]{\includegraphics[width=0.33\textwidth]{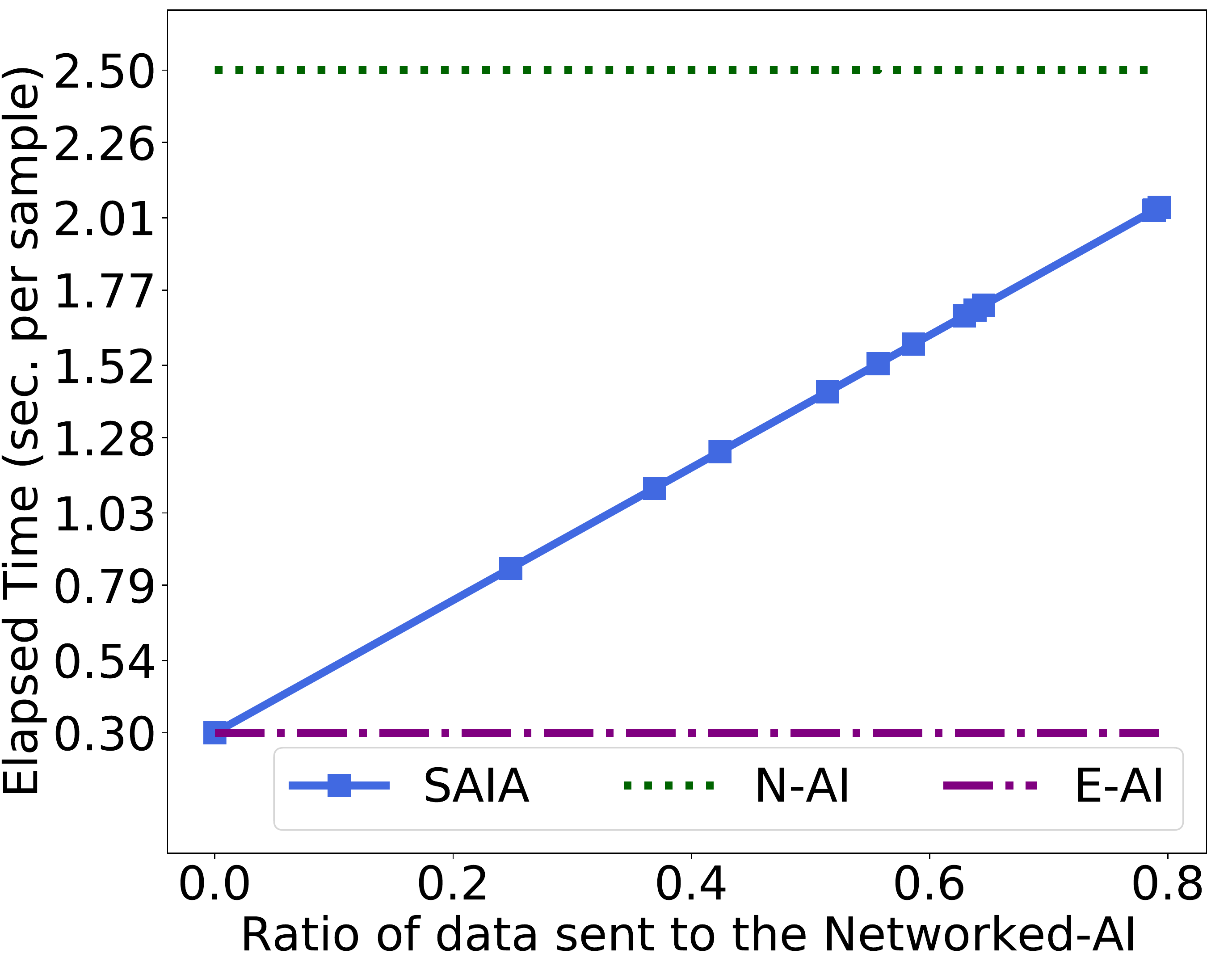}
     \label{fig:eff_fungus_a}}
     \subfloat[]{\includegraphics[width=0.33\textwidth]{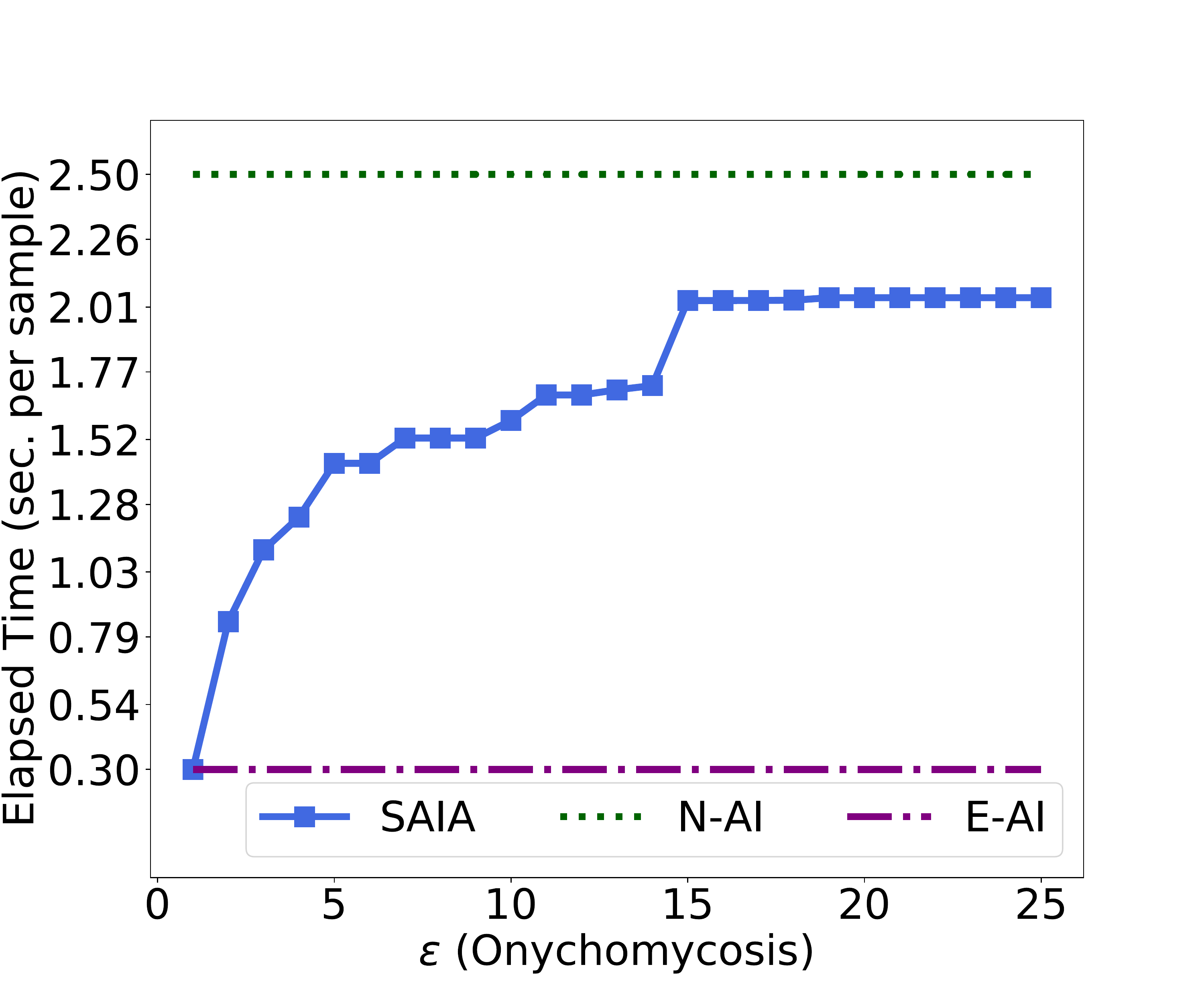}
     \label{fig:eff_fungus_b}}
     \subfloat[]{\includegraphics[width=0.33\textwidth]{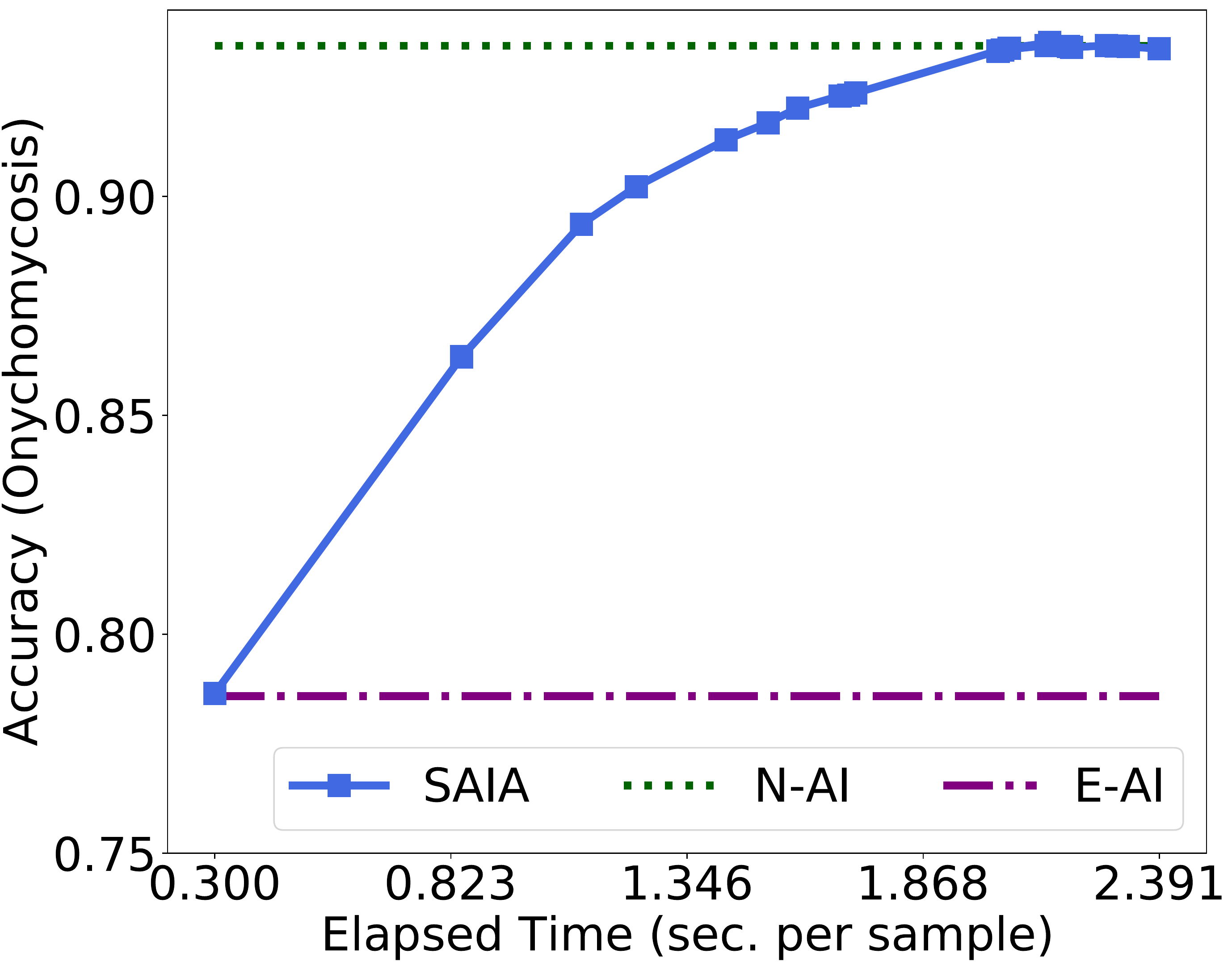}
     \label{fig:eff_fungus_c}}
      \caption{Efficiency Analysis of Onychomycosis Classification}
  \label{fig:eff_fungus}
\end{figure*}

In this section, we evaluate the performance (i.e., accuracy) of our proposed decision unit (i.e., a lightweight binary classifier), and the effectiveness of the hyperparameter $\epsilon$ while influencing the performance of the decision unit. Fig.~\ref{fig:cfm_isic} and Fig.~\ref{fig:cfm_fungus} illustrate the confusion matrices of the decision units while applying different $\epsilon$ for the skin lesion and onychomycosis datasets respectively. As we increase $\epsilon$, more samples that suppose to be sent to the server are determined by the decision unit to be sent to the server (i.e. the true positive rate increases). For instance, in Fig.~\ref{fig:cfm_isic}, as $\epsilon$ increasing from $3$ to $10$, the ratio of data supposed to be sent to the server being sent to the server changed from 0.63 to 0.88. On the other hand, as $\epsilon$ increasing, more samples that suppose to be kept on the client are also determined by the decision unit to be sent to the server (i.e. the false positive rate also increases). For instance, in Fig.~\ref{fig:cfm_isic}, as $\epsilon$ increasing from $3$ to $10$, the ratio of data supposed to be kept on the client being sent to the server changed from 0.26 to 0.44. Fig.~\ref{fig:cfm_fungus} presents the same patterns. However, as illustrated in Fig.~\ref{fig:eff_skin} and Fig.~\ref{fig:eff_fungus}, as we increase $\epsilon$, even though both TPR and FPR are increasing, the TPR is always much higher than the FPR, and the average increasing speed of the TPR is always higher than that of the FPR.

% To summarize,

\subsection{Efficiency Analysis}
In this section, we evaluate the efficiency of our proposed SAIA regarding to the elapsed time averaged over each sample. As illustrated in Fig.~\ref{fig:eff_skin} and Fig~\ref{fig:eff_fungus}, in our experiments on the skin lesion dataset, the elapsed time (second per sample) of the embedded AI and the networked AI are 0.308s and 2.51s respectively, and that of the onychomycosis dataset are 0.3s and 2.5s respectively. Fig.~\ref{fig:eff_skin_a} and Fig~\ref{fig:eff_fungus_a} show that the elapsed time of SAIA system is linearly dependent on the percentage of data sent from the client to the server. Fig.~\ref{fig:eff_skin_b} and Fig~\ref{fig:eff_fungus_b} illustrate that as we increase $\epsilon$, the elapsed time (second per sample) of SAIA would quickly converge to a constant value (e.g., for the skin lesion dataset, while $\epsilon=50$, the elapsed time of SAIA converges to around 1.89s; for the onychomycosis dataset, while $\epsilon=17$, the elapsed time of SAIA converges to around 2.11s). Furthermore, as presented in Fig.~\ref{fig:eff_skin_c} and Fig~\ref{fig:eff_fungus_c}, while SAIA reach the same accuracy as the networked AI, SAIA has less elapsed times on both datasets (i.e., 1.89s vs. 2.51s on the skin lesion dataset, and 2.11s vs. 2.5 on the onychomycosis dataset). To  summarize, even with enough communication resource, by applying the decision unit, our system does not have to send all the data to the server-side, while obtaining the same accuracy as sending all the data to the server-side, and much less processing time compared with the networked AI.

\section{Related Work} \label{sec:relatedWork}
\subsection{Compact Deep Neural Networks} \label{sec:relatedWork_DL_Mobile}
Many real-world applications (e.g., mobile healthcare, smart home, wearable technologies) require to collect and analyze the data on mobile and IoT devices. Hence, compact DNNs have been proposed to conduct inference on such devices. For instance, SqueezeNet \cite{iandola2016squeezenet} obtains AlexNet \cite{krizhevsky2012imagenet} level of accuracy with 50x fewer parameters and less than 0.5MB model size, by downsampling the data using $1\times1$ convolution filters. MobileNet \cite{howard2017mobilenets, sandler2018mobilenetv2, howard2019searching} proposes a useful building block, ``inverted residual block'' into its design of DNNs, that significantly reduces computation complexity without accuracy loss, compared with traditional DNN models. YOLO, a state-of-the-art, real-time object detection system, is designed by using customized architecture, that only has one forth operations of VGG16 \cite{simonyan2014very}.
EfficientNet \cite{tan2019efficientnet} is one of the state-of-the-art DNN models recently proposed for execution on mobile and IoT devices, that uniformly scales each dimension (e.g., width, depth and resolution) of DNN models with a fixed set of scaling coefficients.
Although the compact DNNs could dramatically reduce it computation complexity, the overall performance of compact DNN model still would not be as good as the more advanced models deployed on the server side, that could also be the ensemble/fusion of several well-trained DNN models

\subsection{Compressed Deep Neural Networks} \label{sec:relatedWork_Compressed_DNN}
DNN model compression techniques \cite{hinton2015distilling, ba2014deep, polino2018model, huynh2017deepmon, han2016eie, han2015deep, liu2018demand, zhao2018deepthings} have been proposed to reduce the size and computation workload of DNN models running on the mobile and IoT devices. For instance, Knowledge distillation \cite{hinton2015distilling, ba2014deep, polino2018model} has been proposed to compress a model by teaching a simplified student DNN model, step by step, exactly what to do using a complex pre-trained teacher DNN model, and then deploy the student DNN model on the mobile devices. Network pruning \cite{luo2017thinet} has been proposed to trim the network connections within DNNs that have less influence on the inference accuracy. Data quantization \cite{han2015deep} has been proposed to educe the number of bits to represent each weight value of DNN models.
However, certain recent DNN models, such as MobileNet \cite{howard2019searching} and EfficientNet \cite{tan2019efficientnet} are already very compact and hard to compress significantly. Deploying compressed DNN models on the mobile and IoT devices also cannot take advantage of the more advanced models deployed on the server side.

\subsection{Split Deep Neural Networks} \label{sec:relatedWork_Split_DNN}
Split-DNN architectures \cite{jeong2018computation, kang2017neurosurgeon, lane2016deepx, osia2020hybrid} have been proposed to offload the execution of complex DNN models to compute-capable servers from the mobile or IoT devices, where a DNN is split into head and tail sections, deployed at the client side and the server side, respectively. For instance, Osia et al. \cite{osia2020hybrid} proposes a hybrid architecture where a DNN model, that has previously been trained and fine-tuned on the cloud, would be split into two smaller neural networks: a feature extraction network that runs on the mobile or IoT devices, and a classification network that runs on the cloud system, and both neural networks on the local device and the cloud system would collaborate on running the original complex DNN model.
Matsubara et al. \cite{matsubara2019distilled} proposes a KD-based Split-DNN framework to reduce the communication cost between the client and the server. However, such approaches usually cannot fully rely on the client-side model, thus unable to work if the communication is impeded.
However, none of such approaches directly address the communication bottleneck between the client and the server.  Also, the existing approaches cannot adjust the AI usage on between the client and the serve depending on the device's condition (e.g., storage  size,  power  consumption and  communication bandwidth).

\section{Conclusion} \label{sec:conclusion}
In this paper, we propose SAIA, a novel, effective and efficient split artificial intelligence architecture for mobile healthcare systems, where we design four components: the data pre-processing interface (including objection detection, semantic segmentation and feature extraction), the embedded AI that contains a lightweight classification classifier(s), the networked AI that trains a multi-classifier fusion of several advanced DNN classifiers, and the core component, the decision unit that is another lightweight ML classifier that trains on a set of labeled meta data. %Our proposed meta-information based decision unit could effectively tune whether a sample captured by the client should be operated by embedded AI or networked AI, under different conditions.
A comprehensive experimental evaluation on two large scale healthcare datasets has been conducted. Our results show that SAIA consistently outperforms its baselines in terms of both effectiveness and efficiency. Our proposed decision unit with hyperparameter $\epsilon$ could effectively tune whether a sample captured by the client should be operated by the embedded AI or the networked AI, under different conditions. In our future work, we plan to design and implement fully-fledged split AI architecture that considers more factors, such as energy consumption, communication bandwidth and accuracy requirements.

\section*{Acknowledgments} Effort sponsored in whole or in part by United States Special Operations Command (USSOCOM), under Partnership Intermediary Agreement No. H92222-15-3-0001-01. The U.S. Government is authorized to reproduce and distribute reprints for Government purposes notwithstanding any copyright notation thereon. \footnote{The views and conclusions contained herein are those of the authors and should not be interpreted as necessarily representing the official policies or endorsements, either expressed or implied, of the United States Special Operations Command.}

\bibliography{mybibfile}

\end{document}